# Searching for Bayesian Network Structures in the Space of Restricted Acyclic Partially Directed Graphs

**Silvia Acid**                                                        ACID@DECSAI.UGR.ES
**Luis M. de Campos**                                             LCI@DECSAI.UGR.ES
*Departamento de Ciencias de la Computación e I.A.*
*E.T.S.I. Informática, Universidad de Granada*
*18071 - Granada, SPAIN*

## Abstract

Although many algorithms have been designed to construct Bayesian network structures using different approaches and principles, they all employ only two methods: those based on independence criteria, and those based on a scoring function and a search procedure (although some methods combine the two). Within the score+search paradigm, the dominant approach uses local search methods in the space of directed acyclic graphs (DAGs), where the usual choices for defining the elementary modifications (local changes) that can be applied are arc addition, arc deletion, and arc reversal. In this paper, we propose a new local search method that uses a different search space, and which takes account of the concept of equivalence between network structures: restricted acyclic partially directed graphs (RPDAGs). In this way, the number of different configurations of the search space is reduced, thus improving efficiency. Moreover, although the final result must necessarily be a local optimum given the nature of the search method, the topology of the new search space, which avoids making early decisions about the directions of the arcs, may help to find better local optima than those obtained by searching in the DAG space. Detailed results of the evaluation of the proposed search method on several test problems, including the well-known Alarm Monitoring System, are also presented.

## 1. Introduction

Nowadays, the usefulness of *Bayesian networks* (Pearl, 1988) in representing knowledge with uncertainty and efficient reasoning is widely accepted. A Bayesian network consists of a qualitative part, a *directed acyclic graph* (DAG), and a quantitative one, a collection of numerical parameters, usually conditional probability tables. The knowledge represented in the graphical component is expressed in terms of dependence and independence relationships between variables. These relationships are encoded using the presence or absence of links between nodes in the graph. The knowledge represented in the numerical part quantifies the dependences encoded in the graph, and allows us to introduce uncertainty into the model. All in all, Bayesian networks provide a very intuitive graphical tool for representing available knowledge.

Another attraction of Bayesian networks is their ability to efficiently perform reasoning tasks (Jensen, 1996; Pearl, 1988). The independences represented in the DAG are the key to this ability, reducing changes in the knowledge state to local computations. In addition, important savings in storage requirements are possible since independences allow a factorization of the global numerical representation (the joint probability distribution).

There has been a lot of work in recent years on the automatic learning of Bayesian networks from data. Consequently, there are a great many learning algorithms which may be subdivided into two general approaches: methods based on *conditional independence tests* (also called *constraint-based*), and methods based on a *scoring function*[1] and a *search* procedure.

The algorithms based on independence tests perform a qualitative study of the dependence and independence relationships among the variables in the domain, and attempt to find a network that represents these relationships as far as possible. They therefore take a list of conditional independence relationships (obtained from the data by means of conditional independence tests) as the input, and generate a network that represents most of these relationships. The computational cost of these algorithms is mainly due to the number and complexity of such tests, which can also cause unreliable results. Some of the algorithms based on this approach obtain simplified models (de Campos, 1998; de Campos & Huete, 1997; Geiger, Paz & Pearl, 1990, 1993; Huete & de Campos, 1993), whereas other are designed for general DAGs (de Campos & Huete, 2000a; Cheng, Bell & Liu, 1997; Meek, 1995; Spirtes, Glymour & Scheines, 1993; Verma & Pearl, 1990; Wermuth & Lauritzen, 1983).

The algorithms based on a scoring function attempt to find a graph that maximizes the selected score; the scoring function is usually defined as a measure of fit between the graph and the data. All of them use a scoring function in combination with a search method in order to measure the goodness of each explored structure from the space of feasible solutions. During the exploration process, the scoring function is applied in order to evaluate the fitness of each candidate structure to the data. Each algorithm is characterized by the specific scoring function and search procedure used. The scoring functions are based on different principles, such as entropy (Herskovits & Cooper, 1990; Chow & Liu, 1968; de Campos, 1998; Rebane & Pearl, 1987), Bayesian approaches (Buntine, 1994, 1996; Cooper & Herskovits, 1992; Friedman & Koller, 2000; Friedman, Nachman & Peér, 1999; Geiger & Heckerman, 1995; Heckerman, 1996; Heckerman, Geiger & Chickering, 1995; Madigan & Raftery, 1994; Ramoni & Sebastiani, 1997; Steck, 2000), or the Minimum Description Length (Bouckaert, 1993; Friedman & Goldszmidt, 1996; Lam & Bacchus, 1994; Suzuki, 1993, 1996; Tian, 2000).

There are also hybrid algorithms that use a combination of constraint-based and scoring-based methods: In several works (Singh & Valtorta, 1993, 1995; Spirtes & Meek, 1995; Dash & Druzdzel, 1999; de Campos, Fernández-Luna & Puerta, 2003) the independence-based and scoring-based algorithms are maintained as separate processes, which are combined in some way, whereas the hybridization proposed by Acid and de Campos (2000, 2001) is based on the development of a scoring function that quantifies the discrepancies between the independences displayed by the candidate network and the database, and the search process is limited by the results of some independence tests.

In this paper, we focus on the scoring+search approach. Although algorithms in this category have commonly used local search methods (Buntine, 1991; Cooper & Herskovits, 1992; Chickering, Geiger & Heckerman, 1995; de Campos et al., 2003; Heckerman et al., 1995), due to the exponentially large size of the search space, there is a growing interest in other heuristic search methods, i.e. simulated annealing (Chickering et al., 1995), tabu search (Bouckaert, 1995; Muntenau & Cau, 2000), branch and bound (Tian, 2000), genetic algorithms and evolutionary programming (Larrañaga, Poza, Yurramendi, Murga & Kuijpers, 1996; Myers, Laskey & Levitt, 1999; Wong, Lam & Leung, 1999), Markov chain Monte Carlo (Kocka & Castelo, 2001; Myers et al., 1999), variable neighborhood search (de Campos & Puerta, 2001a; Puerta, 2001), ant colony optimization (de Campos, Fernández-Luna, Gámez & Puerta, 2002; Puerta, 2001), greedy randomized adaptive search procedures (de Campos, Fernández-Luna & Puerta, 2002), and estimation of distribution algorithms (Blanco, Inza & Larrañaga, 2003).

---

1. Some authors also use the term *scoring metric*.

All of these employ different search methods but the same search space: the space of DAGs. A possible alternative is the space of the orderings of the variables (de Campos, Gámez & Puerta, 2002; de Campos & Huete, 2000b; de Campos & Puerta, 2001b; Friedman & Koller, 2000; Larrañaga, Kuijpers & Murga, 1996). In this paper, however, we are more interested in the space of equivalence classes of DAGs (Pearl & Verma, 1990), i.e. classes of DAGs with each representing a different set of probability distributions. There is also a number of learning algorithms that carry out the search in this space (Andersson, Madigan & Perlman, 1997; Chickering, 1996; Dash & Druzdzel, 1999; Madigan, Anderson, Perlman & Volinsky, 1996; Spirtes & Meek, 1995). This feature reduces the size of the search space, although recent results (Gillispie & Perlman, 2001) confirm that this reduction is not as important in terms of the DAG space as previously hoped (the ratio of the number of DAGs to the number of equivalence classes is lower than four). The price we have to pay for this reduction is that the evaluation of the candidate structures does not take advantage of an important property of many scoring functions, namely decomposability, and therefore the corresponding algorithms are less efficient.

In this paper we propose a new search space which is closely related to the space of equivalence classes of DAGs, and which we have called the space of *restricted acyclic partially directed graphs* (RPDAGs). We define a local search algorithm in this space, and show that by using a decomposable scoring function, we can evaluate locally the score of the structures in the neighborhood of the current RPDAG, thus obtaining an efficient algorithm while retaining many of the advantages of using equivalence classes of DAGs. After the original submission of this paper, Chickering (2002) proposed another learning algorithm that searches in the space of equivalence classes of DAGs and which can also score the candidate structures locally, using a canonical representation scheme for equivalence classes, called *completed acyclic partially directed graphs* (CPDAGs).

The rest of the paper is organized as follows: section 2 discusses some preliminaries and the advantages and disadvantages of carrying out the search process in the spaces of DAGs and equivalence classes of DAGs. Section 3 describes the graphical objects, RPDAGs, that will be included in the proposed search space. In Section 4, a detailed description of the local search method used to explore this space is provided. Section 5 shows how we can evaluate RPDAGs efficiently using a decomposable scoring function. Section 6 contains the experimental results of the evaluation of the proposed algorithm on the Alarm (Beinlich, Suermondt, Chavez & Cooper, 1989), Insurance (Binder, Koller, Russell & Kanazawa, 1997) and Hailfinder (Abramson, Brown, Murphy & Winkler, 1996) networks, as well as on databases from the UCI Machine Learning Repository. We also include an empirical comparison with other state-of-the-art learning algorithms. Finally, Section 7 contains the concluding remarks and some proposals for future work.

## 2. DAGs and Equivalence Classes of DAGs

The search procedures used within Bayesian network learning algorithms usually operate on the space of DAGs. In this context, the problem can be formally expressed as: Given a complete training set (i.e. we do not consider missing values or latent variables) $D = \{\mathbf{u}^1, \ldots, \mathbf{u}^m\}$ of instances of a finite set of $n$ variables, $\mathcal{U}$, find the DAG $H^*$ such that

$$H^* = \arg \max_{H \in \mathcal{H}_n} g(H : D) \tag{1}$$

where $g(H : D)$ is a scoring function measuring the fitness of any candidate DAG $H$ to the dataset $D$, and $\mathcal{H}_n$ is the family of all the DAGs with $n$ nodes[2].

Many of the search procedures, including the commonly used local search methods, rely on a neighborhood structure that defines the local rules (operators) used to move within the search space. The standard neighborhood in the space of DAGs uses the operators of arc addition, arc deletion and arc reversal, thereby avoiding (in the first and the third case) the inclusion of directed cycles in the graph.

The algorithms that search in the space of DAGs using local methods are efficient mainly because of the decomposability property that many scoring functions exhibit. A scoring function $g$ is said to be *decomposable* if the score of any Bayesian network structure may be expressed as a product (or a sum in the log-space) of local scores involving only one node and its parents:

$$g(H : D) = \sum_{y \in \mathcal{U}} g_D(y, Pa_H(y)) \tag{2}$$

$$g_D(y, Pa_H(y)) = g(y, Pa_H(y) : N_{y, Pa_H(y)}) \tag{3}$$

where $N_{y, Pa_H(y)}$ are the statistics of the variables $y$ and $Pa_H(y)$ in $D$, i.e. the number of instances in $D$ that match each possible instantiation of $y$ and $Pa_H(y)$. $Pa_H(y)$ will denote the parent set of $y$ in the DAG $H$, i.e. $Pa_H(y) = \{t \in \mathcal{U} \mid t \to y \in H\}$.

A procedure that changes one arc at each move can efficiently evaluate the improvement obtained by this change. Such a procedure can reuse the computations carried out at previous stages, and only the statistics corresponding to the variables whose parent sets have been modified need to be recomputed. The addition or deletion of an arc $x \to y$ in a DAG $H$ can therefore be evaluated by computing only one new local score, $g_D(y, Pa_{H \cup \{x\}}(y))$ or $g_D(y, Pa_{H \setminus \{x\}}(y))$, respectively. The evaluation of the reversal of an arc $x \to y$ requires the computation of two new local scores, $g_D(y, Pa_{H \setminus \{x\}}(y))$ and $g_D(x, Pa_{H \cup \{y\}}(x))$.

It should be noted that each structure in the DAG space is not always different from the others in terms of its representation capability: if we interpret the arcs in a DAG as causal interactions between variables, then each DAG represents a different model; however, if we see a DAG as a set of dependence/independence relationships between variables (that permits us to factorize a joint probability distribution), then different DAGs may represent the same model. Even in the case of using a causal interpretation, if we use observation-only data (as opposed to experimental data where some variables may be manipulated), it is quite common for two Bayesian networks to be empirically indistinguishable. When two DAGs $H$ and $H'$ can represent the same set of conditional independence assertions, we say that these DAGs are *equivalent*[3] (Pearl & Verma, 1990), and we denote this as $H \simeq H'$.

---

2. For reasons of simplicity, the set of nodes which are in one-to-one correspondence with the variables in $\mathcal{U}$ will also be denoted by $\mathcal{U}$.

When learning Bayesian networks from data using scoring functions, two different (but equivalent) DAGs may be indistinguishable, due to the existence of invariant properties on equivalent DAGs, yielding equal scores. We could take advantage of this in order to get a more reduced space of structures to be explored.

The following theorem provides a graphical criterion for determining the equivalence of two DAGs:

**Theorem 1 (Pearl & Verma, 1990)** *Two DAGs are equivalent if and only if they have the same skeleton and the same v-structures.*

The *skeleton* of a DAG is the undirected graph that results from ignoring the directionality of every edge. A *v-structure* in a DAG $H$ is an ordered triplet of nodes, $(x, y, z)$, such that (1) $H$ contains the arcs $x \rightarrow y$ and $y \leftarrow z$, and (2) the nodes $x$ and $z$ are not adjacent in $H$. A *head-to-head pattern* (shortened h-h) in a DAG $H$ is an ordered triplet of nodes, $(x, y, z)$, such that $H$ contains the arcs $x \rightarrow y$ and $y \leftarrow z$. Note that in an h-h pattern $(x, y, z)$ the nodes $x$ and $z$ can be adjacent.

Another characterization of equivalent DAGs was presented by Chickering (1995), together with proof that several scoring functions used for learning Bayesian networks from data give the same score to equivalent structures (such functions are called *score equivalent functions*).

The concept of equivalence of DAGs partitions the space of DAGs into a set of equivalence classes. Whenever a score equivalent function is used, it seems natural to search for the best configuration in this new space of equivalence classes of DAGs. This change in the search space may bring several advantages:

- The space of equivalent classes is more reduced than the space of DAGs (although it is still enormous). We could therefore expect to obtain better results (with the same search effort).

- As we do not spend time generating (using the operators defined to move between neighboring configurations in the search space) and evaluating (using the scoring function) equivalent DAGs, we could obtain more efficient algorithms. However, as the ratio of the number of equivalence classes to the number of DAGs seems (empirically) to asymptote to 0.267 (Gillispie & Perlman, 2001), the efficiency improvement may be small.

- The search in the space of DAGs may be easily trapped in a local optimum, and the situation worsens as the operators defined for this space can move between configurations corresponding to equivalent DAGs (which will be evaluated with the same score). This difficulty can be partially avoided if we search in the space of equivalence classes.

---

3. Several authors also use the term *independence equivalent*, and reserve the term *distribution equivalent* (wrt some family of distributions $\mathcal{F}$) for the more restricted case where the two DAGs can represent the same probability distributions. In the common situation where all the variables are discrete and $\mathcal{F}$ is the family of unrestricted multinomial distributions, these two concepts of equivalence coincide.

The disadvantages are that, in this space of equivalence classes, it is more expensive to generate neighboring configurations, because we may be forced to perform some kind of consistency check, in order to ensure that these configurations represent equivalence classes[4]; in addition, the evaluation of the neighboring configurations may also be more expensive if we are not able to take advantage of the decomposability property of the scoring function. Finally, the new search space might introduce new local maxima that are not present in DAG space.

In order to design an exploring process for the space of equivalence classes we could use two distinct approaches: the first consists in considering that an equivalence class is represented by any of its components (in this case, it is necessary to avoid evaluating more than one component per class); and the second consists in using a canonical representation scheme for the classes.

The graphical objects commonly used to represent equivalence classes of DAGs are *acyclic partially directed graphs* (Pearl & Verma, 1990) (known as *PDAGs*). These graphs contain both directed (arcs) and undirected (links) edges, but no directed cycles. Given a PDAG $G$ defined on a finite set of nodes $\mathcal{U}$ and a node $y \in \mathcal{U}$, the following subsets of nodes are defined:

- $Pa_G(y) = \{t \in \mathcal{U} \mid t \to y \in G\}$, the set of *parents* of $y$.

- $Ch_G(y) = \{t \in \mathcal{U} \mid y \to t \in G\}$, the set of *children* of $y$.

- $Ne_G(y) = \{t \in \mathcal{U} \mid y\!-\!t \in G\}$, the set of *neighbors* of $y$.

- $Ad_G(y) = \{t \in \mathcal{U} \mid t \to y \in G, \text{ or } y \to t \in G \text{ or } y\!-\!t \in G\}$, the set of *adjacents* to $y$. Obviously $Ad_G(y) = Pa_G(y) \cup Ch_G(y) \cup Ne_G(y)$.

An arc $x \to y$ in a DAG $H$ is *compelled* if it appears in every DAG belonging to the same equivalence class as $H$. An arc $x \to y$ in $H$ is said to be *reversible* if it is not compelled, i.e. there is a DAG $H'$ equivalent to $H$ that contains the arc $x \leftarrow y$. As every DAG in a particular equivalence class has the same set of compelled and reversible arcs, a canonical representation of an equivalence class is the PDAG consisting of an arc for every compelled arc in the equivalence class, and a link for every reversible arc. This kind of representation has been given several names: *pattern* (Spirtes & Meek, 1995), *completed PDAG* (*CPDAG*) (Chickering, 1996), *essential graph* (Andersson et al., 1997; Dash & Druzdzel, 1999). As a consequence of theorem 1, a completed PDAG possesses an arc $x \to y$ if and only if a triplet of nodes $(x, y, z)$ forms a v-structure or the arc $x \to y$ is required to be directed due to other v-structures (to avoid forming a new v-structure or creating a directed cycle) (see Figure 1).

Note that an arbitrary PDAG does not necessarily represent some equivalence class of DAGs, although there is a one-to-one correspondence between completed PDAGs and equivalence classes of DAGs. Nevertheless, completed PDAGs are considerably more complicated than general PDAGs. A characterization of the specific properties that a PDAG must verify in order to be a completed PDAG was obtained by Andersson et al. (1997):

---

4. Note that the operators of addition and reversal of an arc in the DAG space also need a consistency check, but in this case we simply test the absence of directed cycles.

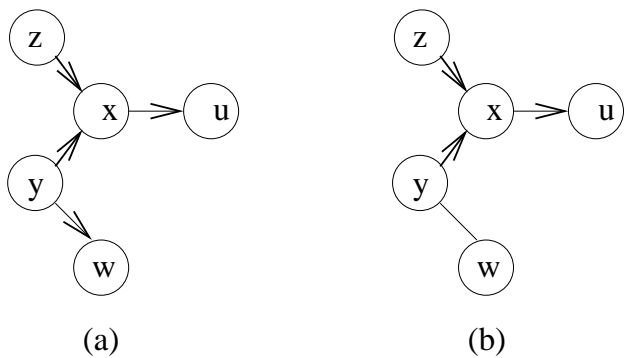

Figure 1: (a) Dag and (b) completed PDAG. The arcs $z \to x$, $y \to x$ and $x \to u$ are compelled; the arc $y \to w$ is reversible

**Theorem 2 (Andersson et al., 1997)** *A PDAG $G$ is a completed PDAG if and only if it satisfies the following conditions:*

1. *$G$ is a chain graph, i.e. it contains no (partially) directed cycles.*

2. *The subgraph induced by every chain component[5] of $G$ is chordal (i.e. on every undirected cycle of length greater than or equal to 4 there are two non-consecutive nodes connected by a link).*

3. *The configuration $x \to y—z$ does not occur as an induced subgraph of $G$.*

4. *Every arc $x \to y \in G$ occurs in at least one of the four configurations displayed in Figure 2 as an induced subgraph of $G$.*

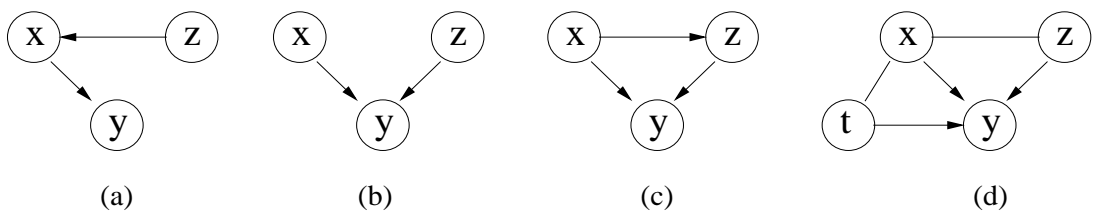

Figure 2: The four different configurations containing an arc $x \to y$ in a completed PDAG

Let us illustrate the advantages of searching in the space of equivalence classes of DAGs rather than the space of DAGs with a simple example. Figure 3 displays the set of possible DAGs involving three nodes $\{x, y, z\}$, with arcs between $z$ and $x$, and between $y$ and $x$. The first three DAGs are equivalent. In terms of independence information, they lead to the same independence statement $I(y, z|x)$ ($y$ and $z$ are conditionally independent given $x$), whereas the statement $I(y, z|\emptyset)$ ($y$ and $z$ are marginally independent) corresponds to the fourth one. The four DAGs may be summarized by only two different completed PDAGs, shown in Figure 4.

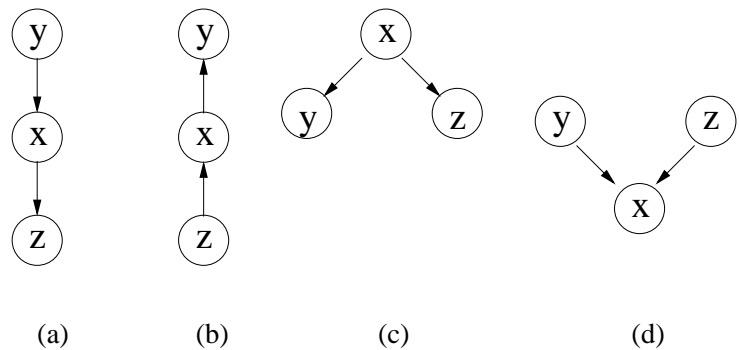

Figure 3: Four different DAGs with three nodes and two arcs

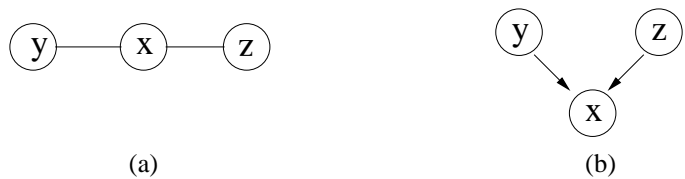

Figure 4: Two different equivalence classes of DAGs

As we can see, the search space may be reduced by using PDAGs to represent the classes: in our example, to two classes instead of four configurations; it can be seen (Andersson et al., 1997) that the ratio of the number of DAGs to the number of classes is 25 / 11 for three nodes, 543 / 185 for four nodes and 29281 / 8792 for five nodes; in more general terms, the results obtained by Gillispie and Perlman (2001) indicate that this ratio approaches a value of less than four as the number of nodes increases. The use of equivalence classes therefore entails convenient savings in exploration and evaluation effort, although the gain is not spectacular.

On the other hand, the use of a canonical representation scheme allows us to explore the space progressively and systematically, without losing any unexplored configuration unnecessarily. Returning to our example, let us suppose that the true model is the DAG displayed in Figure 4.b and we start the search with an empty graph (with no arcs). Let us also assume that the search algorithm identifies that an edge between $x$ and $y$ produces the greatest improvement in the score. At this moment, the two alternatives, $x \to y$ and $x \leftarrow y$ (case 1 and case 2 in Figure 5, respectively), are equivalent. Let us now suppose that we decide to connect the nodes $x$ and $z$; again we have two options: $x \to z$ or $x \leftarrow z$. Nevertheless, depending on the previous selected configuration, we obtain different outcomes that are no longer equivalent (see Figure 5).

If we had chosen case 1 (thus obtaining either case 1.1 or case 1.2), we would have eliminated the possibility of exploring the DAG $z \to x \leftarrow y$, and therefore the exploring process would have been trimmed. As the true model is precisely this DAG (case 2.1 in Figure 5), then the search process would have to include another arc connecting $y$ and $z$

---

5. A chain component of $G$ is any connected component of the undirected graph obtained from $G$ by removing all the arcs.

(cases 1.1.1, 1.2.1 or 1.2.2), because $y$ and $z$ are conditionally dependent given $x$. At this moment, any local search process would stop (in a local optimum), because every local change (arc reversal or arc removal) would make the score worse.

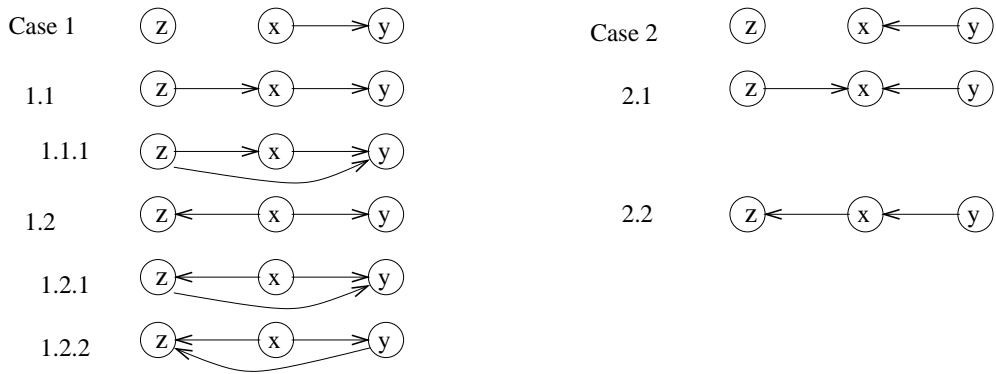

Figure 5: Local search in the space of DAGs is trapped at a local optimum

Consequently, our purpose consists in adding or removing edges (either links or arcs) to the structure without pruning the search space unnecessarily. We could therefore introduce links instead of arcs (when there is not enough information to distinguish between different patterns of arcs), which would serve as templates or dynamic linkers to equivalence patterns. They represent any valid combination of arcs which results in a DAG belonging to the same equivalence class.

Looking again at the previous example, we would proceed as follows: assuming that in our search space the operators of link addition and creation of h-h patterns are available, we would first include the link $x$—$y$; secondly, when considering the inclusion of a connection between $x$ and $z$, we would have two options, shown in Figure 4: the h-h pattern $z \rightarrow x \leftarrow y$ and the pattern $z$—$x$—$y$. In this case the scoring function would assign the greatest score to the h-h pattern $z \rightarrow x \leftarrow y$, thus obtaining the correct DAG.

## 3. Restricted Acyclic Partially Directed Graphs

The scheme of representation that we will use is slightly different from the formalism of completed PDAGs. It is not necessary for each configuration of our search space (which we call *restricted PDAG* or *RPDAG*) to correspond to a different equivalence class; two different RPDAGs may correspond to the same equivalence class. The main reason for this is efficiency: by allowing an equivalence class to be represented (only in some cases) by different RPDAGs, we will gain in efficiency to explore the space. Before explaining this in greater detail, let us define the concept of RPDAG:

**Definition 1 (restricted PDAG)** *A PDAG G is a restricted PDAG (RPDAG) if and only if it satisfies the following conditions:*

*1 $\forall y \in \mathcal{U}$, $Pa_G(y) \neq \emptyset \Rightarrow Ne_G(y) = \emptyset$.*

*2 G does not contain any directed cycle.*

   *3 G does not contain any completely undirected cycle, i.e. a cycle containing only links.*

   *4 If $x \to y$ exists in G then either $|Pa_G(y)| \geq 2$ or $Pa_G(x) \neq \emptyset$.*

     *This condition states that an arc $x \to y$ exists in G only if it is either part of an h-h pattern or there is another arc (originated by an h-h pattern) going to x.*

As an RPDAG is a PDAG, it could be considered to be a representation of a set of (equivalent) DAGs. We therefore must define which the set of DAGs is represented by a given RPDAG $G$, i.e. how direction may be given to the links in $G$ in order to extend it to a DAG. The following definition formalizes this idea.

**Definition 2 (Extension of a PDAG)** *Given a PDAG G, we say that a DAG H is an extension of G if and only if:*

   *1 G and H have the same skeleton.*

   *2 If $x \to y$ is an arc in G then $x \to y$ is also an arc in H (no arc is redirected).*

   *3 G and H have the same h-h patterns (i.e. the process of directing the links in G in order to produce H does not generate new h-h patterns).*

We will use $Ext(G)$ to denote the set of DAGs that are extensions of a given PDAG $G$.

**Proposition 1** *Let G be an RPDAG. Then:*

   *(a) $Ext(G) \neq \emptyset$ (G can be extended to obtain a DAG, i.e. the extension of an RPDAG is well-defined).*

   *(b) $\forall H, H' \in Ext(G)$ $H \simeq H'$ (i.e. all the different DAGs that can be obtained from G by extending it are equivalent).*

**Proof:**
(a) As $G$ has no directed cycle (condition 2 in Definition 1), then either $G$ is already a DAG or it has some links. Let us consider an arbitrary link $x$—$y$. Using condition 1 in Definition 1, neither $x$ nor $y$ can have a parent. We can then direct the link $x$—$y$ in either direction without creating an h-h pattern. If we direct the link $x$—$y$ as $x \to y$ and $y$ is part of another link $y$—$z$, then we direct it as $y \to z$ (in order to avoid a new h-h pattern). We can continue directing the links in a chain in this way, and this process cannot generate a directed cycle because, according to condition 3 in Definition 1, $G$ has no completely undirected cycle.
(b) The extension process of $G$ does not modify the skeleton and does not create new h-h patterns. Therefore, all the extensions of $G$ have the same skeleton and the same v-structures (a v-structure is a particular case of h-h pattern), hence they are equivalent.
□

It should be noted that condition 4 in Definition 1 is not necessary to prove the results in Proposition 1. This condition is included to ensure that the type of PDAG used to represent subsets of equivalent DAGs is as general as possible. In other words, condition 4 guarantees that an RPDAG is a representation of the greatest number of equivalent DAGs, subject to the restrictions imposed by conditions 1-3 in Definition 1. As we will see in the next proposition, this is achieved by directing the minimum number of edges. For example, $z$—$x \to y \to u$ would not be a valid RPDAG. The RPDAG that we would use in this case is $z$—$x$—$y$—$u$.

**Proposition 2** *Let $G$ be a PDAG verifying the conditions 1-3 in Definition 1. There is then a single RPDAG $R$ such that $Ext(G) \subseteq Ext(R)$.*

**Proof:**

The proof is constructive. We shall build the RPDAG $R$ as follows: the skeleton and the h-h patterns of $R$ are the same as those in $G$. An arc $x \to y$ in $G$ shall now be considered such that $Pa_G(x) = \emptyset$ and $Pa_G(y) = \{x\}$ (if such an arc does not exist, then $G$ itself would be an RPDAG): we convert the arc $x \to y$ into the link $x$—$y$. This process is then repeated. Obviously, the PDAG $R$ obtained in this way has no directed cycle and verifies condition 4 in Definition 1. Moreover, we cannot obtain a configuration $z \to x$—$y$ as a subgraph of $R$ because $Pa_G(x) = \emptyset$ (we only remove the direction of arcs whose initial nodes have no parent). In addition, $R$ cannot have any completely undirected cycle because either the arc $x \to y$ is not part of any cycle in $G$ or it is part of a cycle in $G$ that must contain at least one h-h pattern (and the directions of the arcs in this pattern will never be removed). $R$ is therefore an RPDAG.

Let us now prove that $Ext(G) \subseteq Ext(R)$: if $H \in Ext(G)$ then $H$ and $G$ have the same skeleton and h-h patterns, hence $H$ and $R$ also have the same skeleton and h-h patterns. Moreover, as all the arcs in $R$ are also arcs in $G$, if $x \to y \in R$ then $x \to y \in G$, which in turn implies that $x \to y \in H$. Therefore, according to Definition 2, $H \in Ext(G)$.

Finally, let us prove the uniqueness of $R$: we already know that any other RPDAG $R'$ verifying that $Ext(G) \subseteq Ext(R')$ has the same skeleton and h-h patterns as $R$. According to condition 1 in Definition 1, the edges that are not part of any of these h-h patterns but are incident to the middle node $y$ in any h-h pattern $x \to y \leftarrow z$ must be directed away from $y$ (in order to avoid new h-h patterns). The remaining edges that are not part of any h-h pattern must be undirected, in order to satisfy condition 4 in Definition 1. There is therefore only one RPDAG that matches a given skeleton and a set of h-h patterns, so $R$ is the only RPDAG verifying that $Ext(G) \subseteq Ext(R)$. Figure 6 shows an example of the construction process. $\square$

The following proposition ensures that the concept of RPDAG allows us to define a partition in the space of DAGs.

**Proposition 3** *Let $G$ and $G'$ be two different RPDAGs. Then $Ext(G) \cap Ext(G') = \emptyset$.*

**Proof:**

Let $H$ be any DAG. Then $H$ itself is a PDAG and obviously $H = Ext(H)$. By applying the result in Proposition 2, we can assert that there is a single RPDAG $G$ such that $H \subseteq Ext(G)$. $\square$

In the proposition below, we show the properties which are common to all the DAGs belonging to the same extension of an RPDAG.

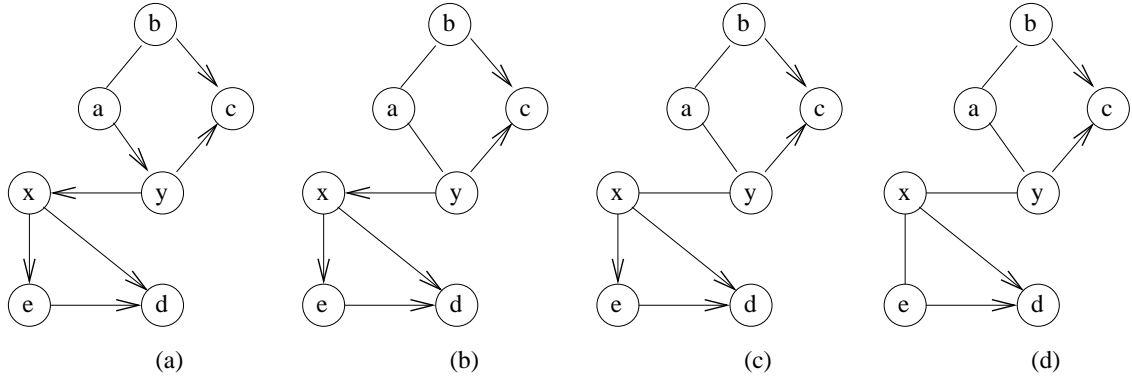

Figure 6: Illustrating the construction process in Proposition 2: (a) PDAG $G$; (b) undirecting the arc $a \to y$; (c) undirecting the arc $y \to x$; (d) undirecting the arc $x \to e$, thus obtaining the RPDAG $R$

**Proposition 4** *Two DAGs belong to the extension of the same RPDAG if and only if they have the same skeleton and the same h-h patterns.*

**Proof:**

The necessary condition is obvious. Let us prove the sufficient condition: let $H$ and $H'$ be two DAGs with common skeleton and h-h patterns. We shall construct a PDAG $G$ as follows: the skeleton and the h-h patterns of $G$ are the same as those in $H$ and $H'$; the edges that have the same orientation in $H$ and $H'$ are directed in $G$ in the same way; the other edges in $G$ remain undirected. From Definition 2, it is clear that $H, H' \in Ext(G)$.

$G$ has no directed cycles because $H$ and $H'$ are DAGs. $G$ has no completely undirected cycles, since all the cycles in $H$ and $H'$ share at least the h-h patterns. In addition, $x \to y\!-\!z$ cannot be a subgraph of $G$ because this would imply the existence of the subgraphs $x \to y \leftarrow z$ and $x \to y \to z$ in $H$ and $H'$, respectively, and therefore these two DAGs would not have the same h-h patterns.

Therefore, the PDAG $G$ satisfies conditions 1-3 in Definition 1. By applying Proposition 2, we can then build a single RPDAG $R$ such that $Ext(G) \subseteq Ext(R)$, hence $H, H' \in Ext(R)$. $\square$

A characterization of the extension of an RPDAG that will be useful later is:

**Proposition 5** *Given an RPDAG $G$ and a DAG $H$, then $H$ is an extension of $G$ if and only if the following conditions hold:*

*1 $G$ and $H$ have the same skeleton.*

*2 $\forall y \in \mathcal{U}$, if $Pa_G(y) \neq \emptyset$ then $Pa_H(y) = Pa_G(y)$.*

*3 $\forall y \in \mathcal{U}$, if $Pa_G(y) = \emptyset$ and $Pa_H(y) \neq \emptyset$ then $|Pa_H(y)| = 1$.*

**Proof:**

• *Necessary condition*:

– $Pa_G(y) \neq \emptyset$: Let $x \in Pa_G(y)$, i.e., $x \rightarrow y \in G$. Then, from condition 2 in Definition 2, $x \rightarrow y \in H$, i.e., $x \in Pa_H(y)$. Moreover, $z \in Pa_G(y) \Leftrightarrow x \rightarrow y \leftarrow z$ is an h-h pattern in $G$. From condition 3 in Definition 2, this occurs if and only if $x \rightarrow y \leftarrow z$ is an h-h pattern in $H$, which is equivalent to $z \in Pa_H(y)$. Therefore, $Pa_H(y) = Pa_G(y)$.

– $Pa_G(y) = \emptyset$ and $Pa_H(y) \neq \emptyset$: Let $x \in Pa_H(y)$. If there is another node $z \in Pa_H(y)$, then $x \rightarrow y \leftarrow z$ is an h-h pattern in $H$ and therefore it is also an h-h pattern in $G$, which contradicts the fact that $Pa_G(y) = \emptyset$. So, $y$ cannot have more than one parent in $H$, hence $|Pa_H(y)| = 1$.

• *Sufficient condition*:
– If $x \rightarrow y \in G$ then $Pa_G(y) \neq \emptyset$. From condition 2 we have $Pa_H(y) = Pa_G(y)$, hence $x \rightarrow y \in H$.
– If $x \rightarrow y \leftarrow z$ is an h-h pattern in $G$, once again from condition 2, $Pa_H(y) = Pa_G(y)$ and therefore $x \rightarrow y \leftarrow z$ is an h-h pattern in $H$.
– If $x \rightarrow y \leftarrow z$ is an h-h pattern in $H$, then $|Pa_H(y)| \neq 1$ and $Pa_H(y) \neq \emptyset$. So, from condition 3, we obtain $Pa_G(y) \neq \emptyset$ and, from condition 2, $Pa_G(y) = Pa_H(y)$. Therefore $x \rightarrow y \leftarrow z$ is an h-h pattern in $G$.  □

## 3.1 Restricted PDAGs and Completed PDAGs

Let us now examine the main differences between the different representations: a representation based on PDAGs ensures that every equivalence class has a unique representation, but there are PDAGs that do not correspond to any equivalence class (in other words, the mapping from equivalence classes to PDAGs is injective). On the other hand, our representation based on RPDAGs guarantees that every RPDAG corresponds to an equivalence class (proposition 1) but does not ensure that every equivalence class has a single representation (the mapping from equivalence classes to RPDAGs is onto). However, the mapping from equivalence classes to CPDAGs is bijective. Figures 7.a, 7.b and 7.c show the three RPDAGs corresponding to the same equivalence class; the associated completed PDAG is shown in Figure 7.d. In this example, the number of DAGs in the equivalence class is 12.

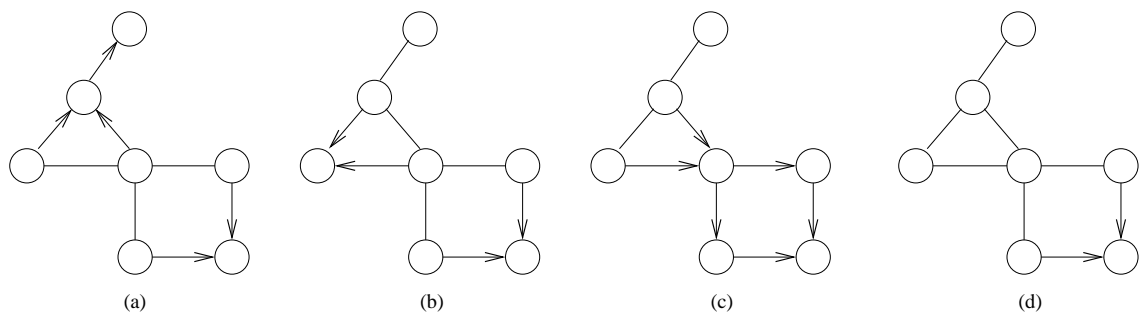

Figure 7: (a), (b) and (c) Three RPDAGs corresponding to the same equivalence class; (d) the associated completed PDAG

As we can see, the difference appears when there are triangular structures. If we compare the definition of RPDAG (Definition 1) with the characterization of CPDAGs (Theorem 2), we may observe that the essential difference is that a CPDAG may have completely undirected cycles, but these cycles must be *chordal*. In RPDAGs, undirected cycles are therefore forbidden, whereas in CPDAGs undirected non-chordal cycles are forbidden.

It should be noted that we could also define RPDAGs by replacing condition 3 in Definition 1 for its equivalent: *The subgraph induced by every chain component of G is a tree* (which is a specific type of chordal graph). In this way, the similarities and differences between CPDAGs and RPDAGs are even clearer. Any of the RPDAGs in the same equivalence class is obtained from the corresponding CPDAG by removing some of the links (converting them into arcs) in order to obtain a tree structure.

Examining the problem from another perspective, from Theorem 1 and Proposition 4 we can see that the role played by the v-structures in CPDAGs is the same as that played by the h-h patterns in RPDAGs.

It is also interesting to note that the number of DAGs which are extensions of a given RPDAG, $G$, can be calculated very easily: the subgraph induced by each chain component of $G$ is a tree, and this tree can be directed in different ways by selecting any of the nodes as the root node. Moreover, we can proceed independently within each chain component. The number of DAGs in $Ext(G)$ is therefore equal to the product of the number of nodes within each chain component of $G$. Regarding the number of RPDAGs that represent the same equivalence class, this number grows exponentially with the size of the undirected cliques in the CPDAG. For example, if the subgraph induced by a chain component in a CPDAG consists of a complete subgraph of $m$ nodes, then the number of RPDAG representations is $\frac{m!}{2}$. This obviously does not mean that a search method based on RPDAGs must explore all these equivalent representations.

Our reason for using RPDAGs is almost exclusively practical. In fact, RPDAGs do not have a clear semantics (they are a somewhat hybrid creature, between DAGs and completed PDAGs). We can only say that RPDAGs would correspond to sets of equivalent DAGs which share all the causal patterns where an effect node has at least two causes (and only the causal patterns where a single cause provokes an effect are not determined). This is not problematic when we are performing model *selection* but it becomes critical if we are doing model *averaging*: without a semantic understanding of the class of RPDAGs, it will be quite difficult to assign a prior to them.

Our intention is to trade the uniqueness of the representation of equivalence classes of DAGs (CPDAGs) for a more manageable one (RPDAGs): testing whether a given PDAG $G$ is an RPDAG is easier than testing whether $G$ is a completed PDAG. In the first case, the consistency check involves testing for the absence of directed and completely undirected cycles (the complexity of these tests and those necessary to verify whether a directed graph is a DAG is exactly the same), whereas in the second case, in addition to testing for the absence of directed and partially directed cycles, we also need to perform chordality tests and check that each arc is part of one of the induced subgraphs displayed in Figure 2. The price we have to pay for using RPDAGs is that we may occasionally need to evaluate an equivalence class more than once. In the next section, we will examine how a local search method which uses RPDAGs can also take advantage of the decomposability property of a scoring function in order to efficiently evaluate neighboring structures.

## 4. The Search Method

We will use a local method to explore the search space of RPDAGs. The starting point of the search process will be an empty RPDAG (corresponding to an empty DAG). Nevertheless, we could start from another configuration if we have some prior knowledge about the presence or absence of some edges or v-structures. We must then define the operators to move from one configuration to another neighboring configuration.

### 4.1 Overview

Our basic operators are the inclusion of an edge between a pair of non-adjacent nodes and the removal of an existing edge between a pair of adjacent nodes in the current configuration. These edges may be either directed or undirected.

The inclusion of an isolated link $x$—$y$ will serve as a template for the arcs $x \rightarrow y$ and $x \leftarrow y$; however, the link $x$—$y$ together with another link $x$—$z$ represent any combination of arcs except those that create new h-h patterns (the DAGs (a), (b) and (c) in Figure 3). In the case of adding an arc, we may obtain several different neighboring configurations, depending on the topology of the current RPDAG and the direction of the arc being included. As we will see, if we are testing the inclusion of an edge between two nodes $x$ and $y$, this may involve testing some of the different valid configurations obtained by the inclusion of the link $x$—$y$, the arc $x \rightarrow y$, the arc $x \leftarrow y$, the h-h pattern $x \rightarrow y \leftarrow z$ or the h-h pattern $z \rightarrow x \leftarrow y$ (where $z$ in the last two cases would be any node such that either the link $y$—$z$ or the link $z$—$x$ exists in the current configuration). However, the removal of an edge will always result in only one neighboring configuration. Other operators, such as arc reversal (Chickering, 1996), will not be used by our search method. The set of neighboring configurations of a given RPDAG $G$ will therefore be the set of all the different RPDAGs obtained from $G$ by adding or deleting a single edge (either directed or undirected).

Before explaining the details of the search method, let us illustrate the main ideas by means of the following example: consider the RPDAG in Figure 8, which represents the current configuration of the search process (this figure only displays the part of the RPDAG corresponding to the neighborhood of the nodes $x$ and $y$), and assume that we shall include an edge between the nodes $x$ and $y$.

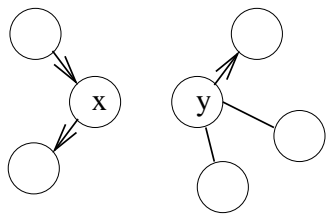

Figure 8: Node $x$ has one parent and one child, $y$ does not have any parents and has two neighbors

In this situation, we cannot introduce the link $x$—$y$ because we would violate one of the conditions defining RPDAGs (condition 1). We may introduce the arc $x \rightarrow y$ and in this case, again in order to preserve condition 1, the two neighbors of $y$ must be converted into children. We can also include the arc $x \leftarrow y$. Finally, we may include two different h-h patterns $x \rightarrow y \leftarrow z$, where $z$ is a neighbor of $y$ (the other neighbor must be converted into a child, once again in order to preserve condition 1). These four different configurations are displayed in Figure 9.

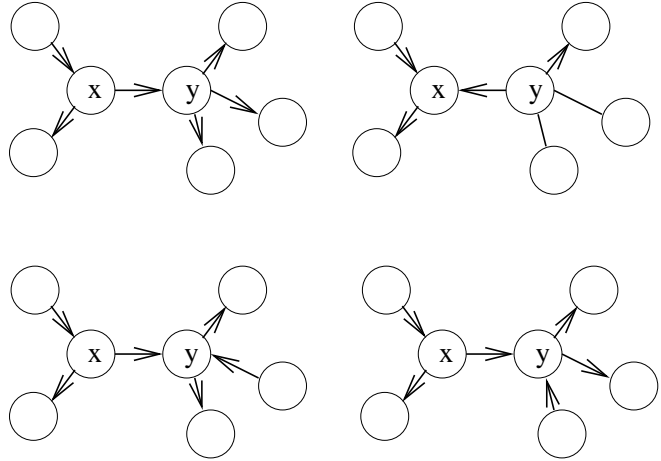

Figure 9: Neighboring configurations obtained by including an edge between $x$ and $y$ in the configuration of Fig. 8

## 4.2 Neighboring Configurations

In order to design a systematic way to determine which neighboring RPDAGs arise from the inclusion or the removal of an edge in an RPDAG, it is sufficient to consider some local parameters of the two nodes to be connected. First, some additional notation is introduced. If $|\cdot|$ represents the cardinality of a set, given a node $x$ in a PDAG $G$, we define:

- $p_G(x) = |Pa_G(x)|$,   •  $c_G(x) = |Ch_G(x)|$

- $n_G(x) = |Ne_G(x)|$,   •  $a_G(x) = |Ad_G(x)|$

Observe that for any RPDAG $G$, the following two properties hold:

- $p_G(x) + c_G(x) + n_G(x) = a_G(x)$

- if $p_G(x) \neq 0 \Rightarrow n_G(x) = 0$ (hence $p_G(x) + c_G(x) = a_G(x)$)

### 4.2.1 Adding Edges

The number and type of neighboring configurations that can be obtained from the inclusion in the current RPDAG of an edge between $x$ and $y$ can be determined from the parameters above. The resultant casuistry may be reduced to seven states, which we have labeled from $A$ to $G$. In order to facilitate its description, we shall use the decision tree shown in Figure 10[6].

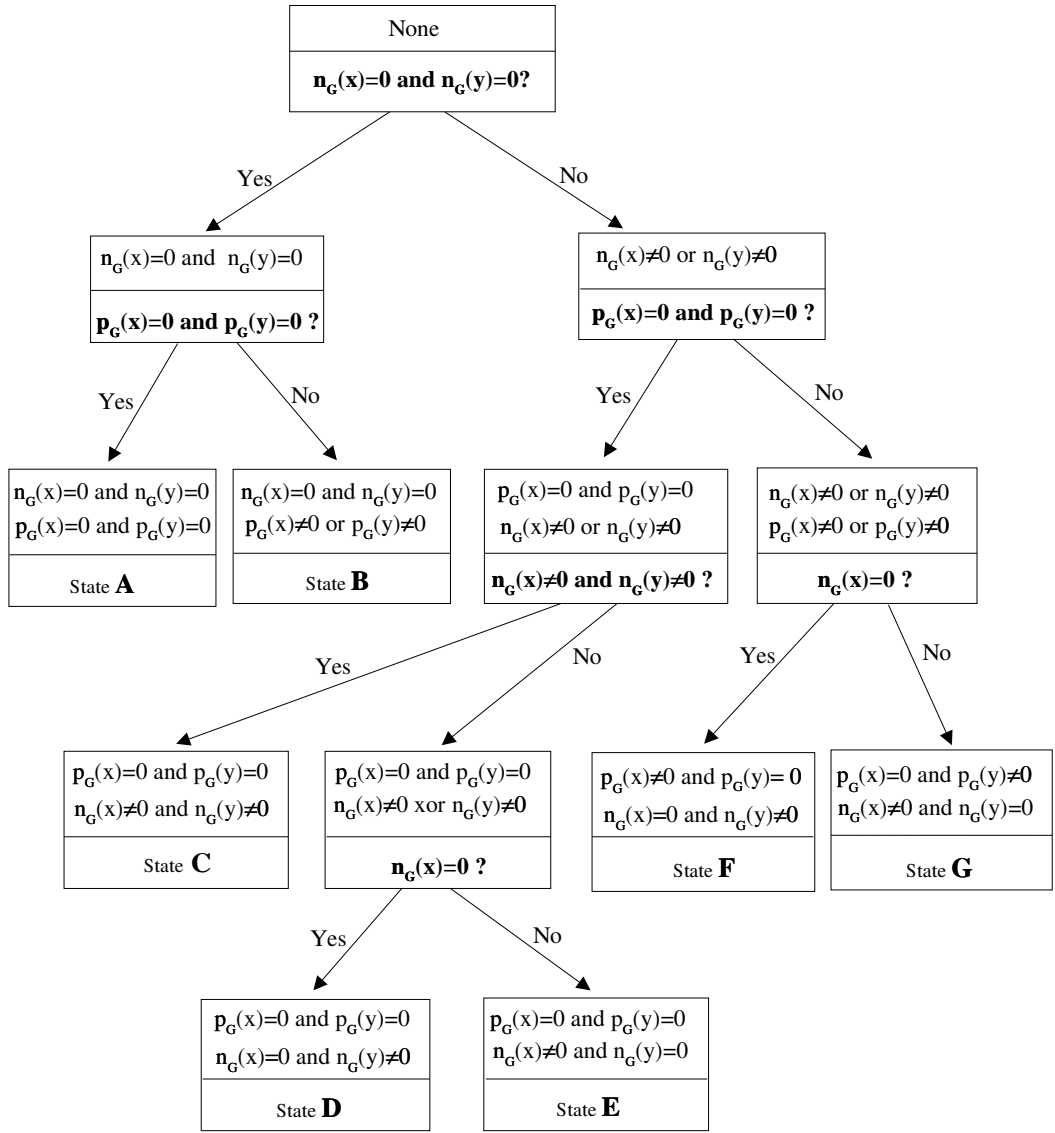

Figure 10: The tree of possible states that may result by adding an edge between nodes $x$ and $y$

---

6. This tree could be organized differently in order to improve the efficiency in the location of the current state. However, this particular tree was selected so as to clarify the presentation.

In this tree, the lower box of each non-terminal vertex contains a test (about the number of parents or the number of neighbors of nodes $x$ and $y$). The lower box of each terminal vertex contains the label of the state resulting from following the path from the root to that terminal vertex. The description of each state (i.e. the different neighboring configurations that can be obtained in this case) can be found in Table 1. The upper boxes of all the vertices in the tree show the restrictions imposed on each intermediate or terminal state. For example, state B corresponds to a situation where both nodes $x$ and $y$ do not have neighbors and at least one of them has some parent. Although the tree has seven different states, there are only five truly different states, since states D and E, and states F and G are symmetrical.

| State | Number of configurations | Added edges | Directed Cycles ? | Undirected Cycles ? | Completing ? |
|---|---|---|---|---|---|
| A | 1 | $x$—$y$ | No | No | No |
| B | 2 | $x \to y$ | Yes[1] | No | No |
|   |   | $x \leftarrow y$ | Yes[2] | No | No |
| C | $n_G(x) + n_G(y) + 1$ | $x$—$y$ | No | Yes | No |
|   |   | $x \to y \leftarrow z$ | Yes[3] | No | Yes[3] |
|   |   | $z \to x \leftarrow y$ | Yes[4] | No | Yes[4] |
| D | $n_G(y) + 1$ | $x$—$y$ | No | No | No |
|   |   | $x \to y \leftarrow z$ | No | No | Yes[3] |
| E | $n_G(x) + 1$ | $x$—$y$ | No | No | No |
|   |   | $z \to x \leftarrow y$ | No | No | Yes[4] |
| F | $n_G(y) + 2$ | $x \leftarrow y$ | No | No | No |
|   |   | $x \to y$ | Yes | No | Yes |
|   |   | $x \to y \leftarrow z$ | Yes[3] | No | Yes[3] |
| G | $n_G(x) + 2$ | $x \to y$ | No | No | No |
|   |   | $x \leftarrow y$ | Yes | No | Yes |
|   |   | $z \to x \leftarrow y$ | Yes[4] | No | Yes[4] |
| (1) only if $p_G(x) \neq 0$ and $c_G(y) \neq 0$ | | | (3) only if $n_G(y) \geq 2$ | | |
| (2) only if $p_G(y) \neq 0$ and $c_G(x) \neq 0$ | | | (4) only if $n_G(x) \geq 2$ | | |

Table 1: Table of states that may result by adding an edge between nodes $x$ and $y$

In Table 1, each row corresponds to a state: the first column contains the labels of the states; the second column displays the total number of neighboring configurations that can be obtained for each state; the third column shows the different types of edges that, for each state, can be added to the current configuration; columns four, five and six will be discussed later.

Using the example in Figure 8, we shall explain the use of the decision tree as well as the instantiation of the information in Table 1. Following the decision tree, at level 1 (the root vertex), the test is false since $y$ has two neighbors. At level 2, the test is also false as $x$ has one parent. At level 3 the test is true, since $x$ has no neighbor. At level 4 we reach a terminal vertex. Our current configuration therefore corresponds to state F. Then, by examining state F in Table 1, we can confirm that we reach four different configurations ($n_G(y) = 2$): $G \cup \{x \leftarrow y\}$, $G \cup \{x \rightarrow y\}$ without new h-h patterns, and two $G \cup \{x \rightarrow y \leftarrow z\}$ which produce new h-h patterns. So, these are the only structures that our algorithm must evaluate when considering the inclusion of the candidate edge $x$—$y$. In Figure 14, we show an example for each of the five non-symmetrical states (once again, these examples only display the part of the RPDAGs corresponding to the neighborhood of the nodes $x$ and $y$).

We therefore have a systematic way to explore all the neighboring configurations which result from adding an edge. However, it will sometimes be necessary to perform some additional steps since the configurations obtained must be RPDAGs:

- First, we must maintain the condition 1 ($p_G(y) \neq 0 \Rightarrow n_G(y) = 0$). It is therefore necessary to *complete* the configuration for some of the described states, i.e. some of the links must be converted into arcs. The completing process consists in firing an orientation in cascade, starting from the links $y$—$t$ such that the arc just introduced is $x \rightarrow y$. Let us consider the situation in Figure 11, where we want to connect the nodes $x$ and $y$, a case corresponding to state D. Among the three possible neighboring configurations, let us suppose that we are testing the one which introduces the h-h pattern $x \rightarrow y \leftarrow z$. The RPDAG obtained from the completing process is also displayed in Figure 11. The sixth column in Table 1 shows which states and neighboring configurations may require the completing process.

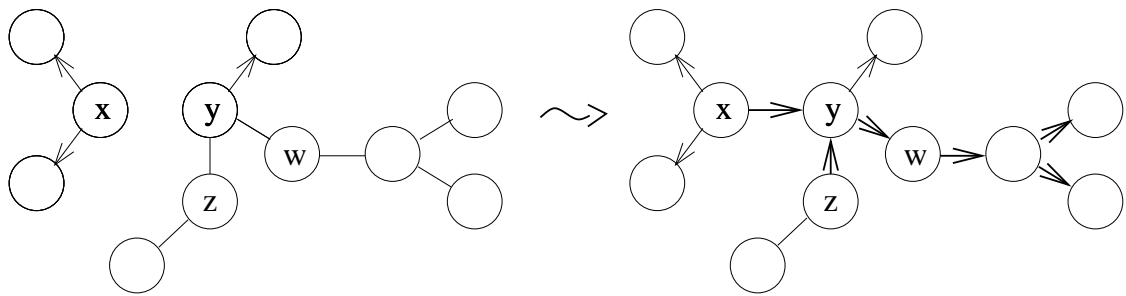

Figure 11: Transformation of a configuration after including the pattern $x \rightarrow y \leftarrow z$ and completing

- Secondly, it is possible that some of the neighboring configurations must be rejected, as they give rise to directed or completely undirected cycles (conditions 2 and 3 defining RPDAGs). For example, let us consider the situation displayed in Figure 12, which corresponds to state F. In this case, the configuration obtained after including the arc $x \rightarrow y$ and completing would generate a directed cycle. This configuration must therefore be rejected. The columns four and five in Table 1 show which states and configurations may require a detection of directed or completely undirected cycles, respectively.

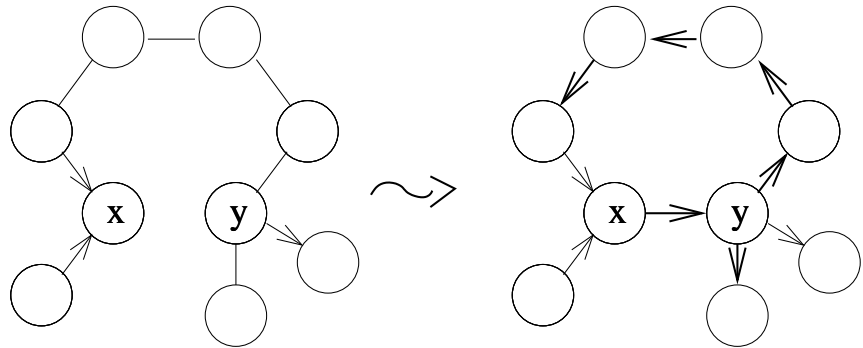

Figure 12: Neighboring configuration that gives rise to a directed cycle

### 4.2.2 DELETING EDGES

The other basic operator, the removal of an edge (either link or arc) is much simpler than the addition of an edge, since only one neighboring configuration is obtained when we delete an edge. Moreover, it is not necessary to perform any test for detecting directed or undirected cycles. However, in this case we need to preserve condition 4 in the definition of RPDAG (if $x \to y$ exists in $G \Rightarrow |Pa_G(y)| \geq 2$ or $Pa_G(x) \neq \emptyset$), that could be violated after an arc is deleted. This situation may appear only when we are going to remove an arc $x \to y$ and either $Pa_G(y) = \{x\}$ or $Pa_G(y) = \{x, u\}$: in the first case, all the children of $y$ that do not have other parents than $y$ must be converted into neighbors of $y$, and this process is repeated starting from each of these children; in the second case, if $Pa_G(u) = \emptyset$ then in addition to the previous process, the arc $u \to y$ must be converted into the link $u$—$y$. Figure 13 illustrates these situations. This procedure of transforming arcs into links is exactly the same as the one described in the proof of Proposition 2.

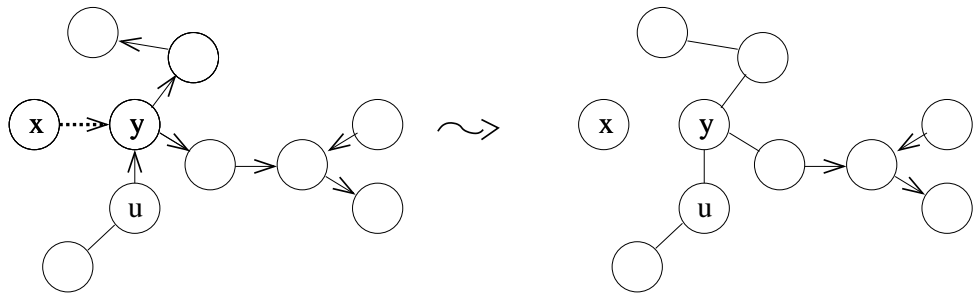

Figure 13: Transforming arcs into links after removing the arc $x \to y$

## 4.3 The Operators

Although the previous description of the operators that define the neighborhood of an RPDAG is quite convenient from a practical (implementation) point of view, for the sake of clarity, we shall describe them in another way. In fact, we shall use five operators:

- $A\_arc(x, y)$, addition of an arc $x \to y$.

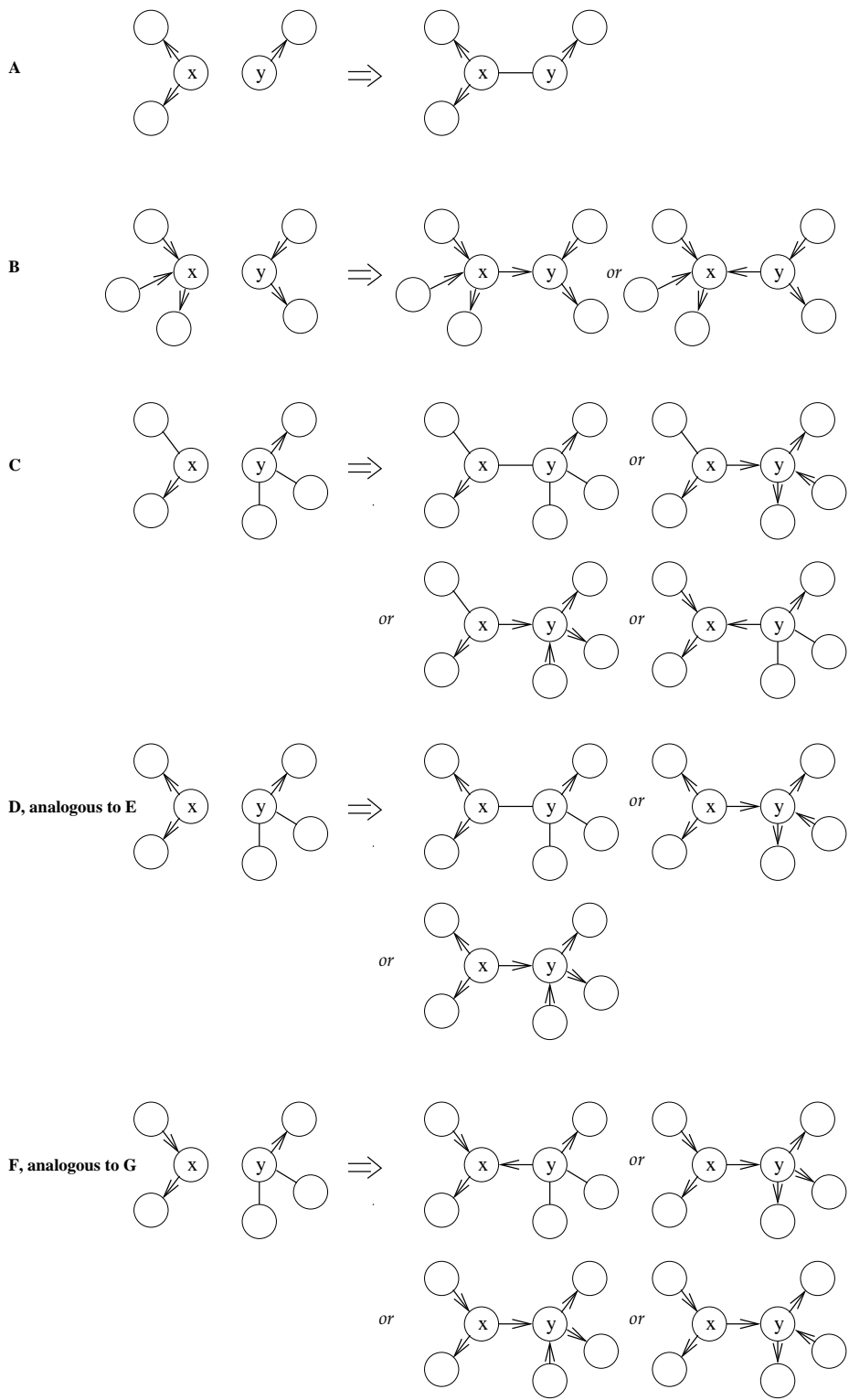

Figure 14: For each state of the decision tree in Fig. 10, an example of the neighboring configurations of an RPDAG that can be obtained after adding an edge between nodes $x$ and $y$

- $A\_link(x, y)$, addition of a link $x$—$y$.

- $D\_arc(x, y)$, deletion of an arc $x \to y$.

- $D\_link(x, y)$, deletion of a link $x$—$y$.

- $A\_hh(x, y, z)$, addition of an arc $x \to y$ and creation of the h-h pattern $x \to y \leftarrow z$ by transforming the link $y$—$z$ into the arc $y \leftarrow z$.

The conditions that the current RPDAG $G$ must verify so that each of these operators can be applied in order to obtain a valid neighboring RPDAG are shown in Table 2. These conditions can be easily derived from the information in Figure 10 and Table 1. In Table 2, $UC(x$—$y)$ represents a test for detecting completely undirected cycles after inserting the link $x$—$y$ in the current RPDAG. Note that we can perform this test very easily without actually inserting the link: it is only necessary to check the existence of a path between $x$ and $y$ exclusively formed by the links. Similarly, $DC(x \to y)$ and $DC(x \to y \leftarrow z)$ represent tests for detecting directed cycles after inserting the arc $x \to y$ and the h-h pattern $x \to y \leftarrow z$ in the current RPDAG, respectively (and perhaps completing). It should also be noted that we can perform these tests without inserting the arc or the h-h pattern: in this case we only need to check the existence of a path from $y$ to $x$ containing only either links or arcs directed away from $y$ (a partially directed path from $y$ to $x$). Table 2 also shows which operators may require a post-processing step in order to ensure that the corresponding neighboring configuration of $G$ is an RPDAG. In Table 2, $Complete(y)$ and $Undo(y)$ refer to the procedures that preserve conditions 1 and 4 in Definition 1, respectively. Note that both $UC(x$—$y)$ and $Complete(y)$ take time $O(l_y)$ in the worst case, where $l_y$ is the number of links in the subgraph induced by the chain component of $G$ that contains $y$; $Undo(y)$ takes time $O(d_y)$ in the worst case, where $d_y$ is the number of arcs in the subgraph induced by the set of descendants of $y$ in $G$ that only have one parent; $DC(x \to y)$ and $DC(x \to y \leftarrow z)$ both take time $O(dl_y)$ in the worst case, where $dl_y$ is the number of edges (either arcs or links) in the subgraph induced by the nodes in the chain component of $G$ that contains $y$ together with their descendants.

## 5. The Exploring Process and the Evaluation of Candidate Structures

The search method we have described may be applied in combination with any score equivalent function $g$ (for example the AIC, BIC, MDL and BDe scoring functions are score equivalent). An easy (but inefficient) way to integrate our search method with a score equivalent function would be as follows: given an RPDAG $G$ to be evaluated, select any extension $H$ of $G$ and compute $g(H : D)$. We could also use other (non-equivalent) scoring functions, although the score of $G$ would depend on the selected extension.

However, let us consider the case of a decomposable scoring function $g$: the DAG obtained by adding or removing an arc from the current DAG $H$ can be evaluated by modifying only one local score:

$$g(H \cup \{x \to y\} : D) = g(H : D) - g_D(y, Pa_H(y)) + g_D(y, Pa_H(y) \cup \{x\})$$
$$g(H \setminus \{x \to y\} : D) = g(H : D) - g_D(y, Pa_H(y)) + g_D(y, Pa_H(y) \setminus \{x\})$$

| Operator | Conditions | Post-processing |
|---|---|---|
| $A\_arc(x,y)$ | $x \notin Ad_G(y)$; $p_G(x) \neq 0$ or $p_G(y) \neq 0$;<br>if $(p_G(x) \neq 0$ and $(c_G(y) \neq 0$ or $n_G(y) \neq 0))$<br>   then $DC(x \rightarrow y) = False$ | if $p_G(x) \neq 0$ and $n_G(y) \neq 0$<br>   then $Complete(y)$ |
| $A\_link(x,y)$ | $x \notin Ad_G(y)$; $p_G(x) = 0$ and $p_G(y) = 0$;<br>if $(n_G(x) \neq 0$ and $n_G(y) \neq 0)$<br>   then $UC(x\!-\!y) = False$ | None |
| $D\_arc(x,y)$ | $x \in Pa_G(y)$ | if $p_G(y) \leq 2$<br>   then $Undo(y)$ |
| $D\_link(x,y)$ | $x \in Ne_G(y)$ | None |
| $A\_hh(x,y,z)$ | $x \notin Ad_G(y)$; $z \in Ne_G(y)$;<br>$p_G(y) = 0$ and $n_G(y) \neq 0$;<br>if $(n_G(y) \geq 2$ and $(p_G(x) \neq 0$ or $n_G(x) \neq 0))$<br>   then $DC(x \rightarrow y \leftarrow z) = False$ | if $n_G(y) \geq 2$<br>   then $Complete(y)$ |

Table 2: The operators, their conditions of applicability, and post-processing requirements

Using decomposable scoring functions, the process of selecting, given an RPDAG, a representative DAG and then evaluating it may be quite inefficient, since we would have to recompute the local scores for all the nodes instead of only one local score. This fact can make a learning algorithm that searches in the space of equivalence classes of DAGs considerably slower than an algorithm that searches in the space of DAGs (this is the case of the algorithm proposed by Chickering, 1996).

Our search method can be used for decomposable scoring functions so that: (1) it is not necessary to transform the RPDAG into a DAG, the RPDAG can be evaluated directly, and (2) the score of any neighboring RPDAG can be obtained by computing at most two local scores. All the advantages of the search methods on the space of DAGs are therefore retained, but a more reduced and robust search space is used.

Before these assertions are proved, let us examine an example. Consider the RPDAG $G$ in Figure 15 and the three neighboring configurations produced by the inclusion of an edge between $x$ and $y$, $G_1$, $G_2$ and $G_3$ (also displayed in Figure 15).

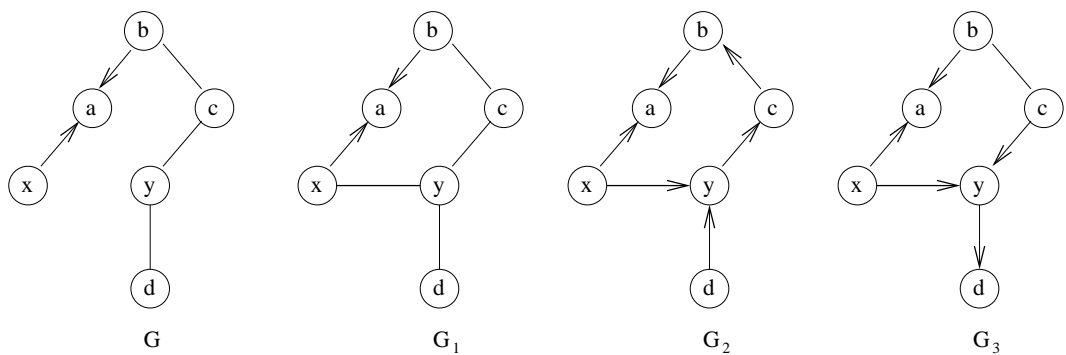

Figure 15: An RPDAG $G$ and three neighboring configurations $G_1$, $G_2$ and $G_3$

The score of each of these RPDAGs is equal to the score of any of their extensions. Figure 16 displays one extension for each neighboring configuration.

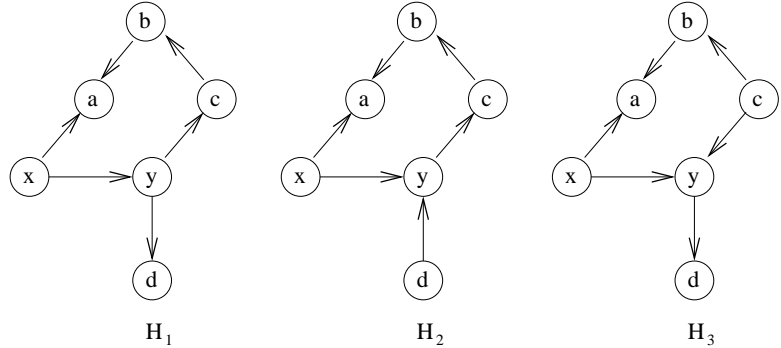

Figure 16: Extensions $H_1$, $H_2$ and $H_3$ of the RPDAGs $G_1$, $G_2$ and $G_3$ in Fig. 15

We can therefore write:

$$g(G_1 : D) = g(H_1 : D) = g_D(x, \emptyset) + g_D(a, \{xb\}) + g_D(b, c) + g_D(c, y) + g_D(y, x) + g_D(d, y)$$
$$g(G_2 : D) = g(H_2 : D) = g_D(x, \emptyset) + g_D(a, \{xb\}) + g_D(b, c) + g_D(c, y) + g_D(y, \{xd\}) + g_D(d, \emptyset)$$
$$g(G_3 : D) = g(H_3 : D) = g_D(x, \emptyset) + g_D(a, \{xb\}) + g_D(b, c) + g_D(c, \emptyset) + g_D(y, \{xc\}) + g_D(d, y)$$

For each extension $H_i$ of any neighboring configuration $G_i$, it is always possible to find an extension $H_{Gi}$ of the current RPDAG $G$ such that the scores of $H_i$ and $H_{Gi}$ only differ in one local score (Figure 17 displays these extensions). We can then write:

$$g(G : D) = g(H_{G1} : D) = g_D(x, \emptyset) + g_D(a, \{xb\}) + g_D(b, c) + g_D(c, y) + g_D(y, \emptyset) + g_D(d, y)$$
$$g(G : D) = g(H_{G2} : D) = g_D(x, \emptyset) + g_D(a, \{xb\}) + g_D(b, c) + g_D(c, y) + g_D(y, d) + g_D(d, \emptyset)$$
$$g(G : D) = g(H_{G3} : D) = g_D(x, \emptyset) + g_D(a, \{xb\}) + g_D(b, c) + g_D(c, \emptyset) + g_D(y, c) + g_D(d, y)$$

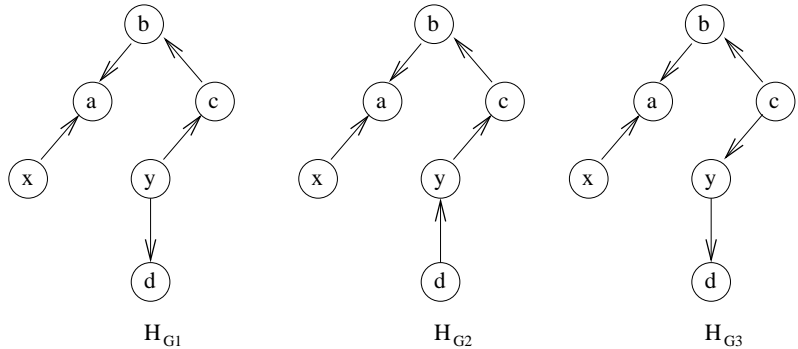

Figure 17: Three different extensions $H_{G1}$, $H_{G2}$ and $H_{G3}$ of the RPDAG $G$ in Fig. 15

Taking into account the previous expressions, we obtain:

$$g(G_1 : D) = g(G : D) - g_D(y, \emptyset) + g_D(y, x)$$

$$g(G_2 : D) = g(G : D) - g_D(y, d) + g_D(y, \{xd\})$$
$$g(G_3 : D) = g(G : D) - g_D(y, c) + g_D(y, \{xc\})$$

Therefore, the score of any neighboring configuration may be obtained from the score of $G$ by computing only two local scores. Note that some of these local scores may have already been computed at previous iterations of the search process: for example, $g_D(y, \emptyset)$ had to be used to score the initial empty RPDAG, and either $g_D(y, d)$ or $g_D(y, c)$ could have been computed when the link $y$—$d$ or $y$—$c$ was inserted into the structure.

**Proposition 6** *Let $G$ be an RPDAG and $G'$ be any RPDAG obtained by applying one of the operators described in Table 2 to $G$. Let $g$ be a score equivalent and decomposable function.*

(a) *If the operator is $A\_link(x, y)$ then*

$$g(G' : D) = g(G : D) - g_D(y, \emptyset) + g_D(y, \{x\})$$

(b) *If the operator is $A\_arc(x, y)$ then*

$$g(G' : D) = g(G : D) - g_D(y, Pa_G(y)) + g_D(y, Pa_G(y) \cup \{x\})$$

(c) *If the operator is $A\_hh(x, y, z)$ then*

$$g(G' : D) = g(G : D) - g_D(y, \{z\}) + g_D(y, \{x, z\})$$

(d) *If the operator is $D\_link(x, y)$ then*

$$g(G' : D) = g(G : D) - g_D(y, \{x\}) + g_D(y, \emptyset)$$

(e) *If the operator is $D\_arc(x, y)$ then*

$$g(G' : D) = g(G : D) - g_D(y, Pa_G(y)) + g_D(y, Pa_G(y) \setminus \{x\})$$

**Proof:**

(1) First, we shall prove that we can construct an extension $H'$ of $G'$ and another extension $H$ of $G$, such that $H$ and $H'$ differ in only one arc (this arc being $x \to y$).

• Consider the cases (a), (b), and (c), which correspond to the addition of an edge between $x$ and $y$: in case (a), $G' = G \cup \{x$—$y\}$ and let $H'$ be an extension of $G'$ that contains the arc $x \to y$; in case (b), where $G' = G \cup \{x \to y\}$, and in case (c), where $G' = (G \setminus \{y$—$z\}) \cup \{x \to y \leftarrow z\}$, let $H'$ be any extension of $G'$ (which will contain the arc $x \to y$). In all three cases, let $H = H' \setminus \{x \to y\}$. We shall prove that $H$ is an extension of $G$:

   – First, it is obvious that $G$ and $H$ have the same skeleton.

   – Secondly, if $u \to v \in G$ (in either case $u \to v \neq x \to y$), then $u \to v \in G'$. As $H'$ is an extension of $G'$, then $u \to v \in H'$, and this implies that $u \to v \in H$. Therefore, all the arcs in $G$ are also arcs in $H$. This result also ensures that every h-h pattern in $G$ is also an h-h pattern in $H$.

– Thirdly, if $u \to v \leftarrow w$ is an h-h pattern in $H$ (in either case $u \to v \leftarrow w \neq x \to y \leftarrow w$), then $u \to v \leftarrow w \in H'$. Once again, as $H'$ is an extension of $G'$, we can see that $u \to v \leftarrow w \in G'$, and then $u \to v \leftarrow w \in G$. So, $G$ and $H$ have the same h-h patterns.

$H$ is therefore an extension of $G$, according to Definition 2. Note that $\forall u \neq y$ $Pa_H(u) = Pa_{H'}(u)$ and $Pa_H(y) = Pa_{H'}(y) \setminus \{x\}$.

• Let us now consider cases (d) and (e), which correspond to the deletion of an edge between $x$ and $y$ (either a link or an arc, respectively): in case (d), let $H$ be an extension of $G$ containing the arc $x \to y$; in case (e), let $H$ be any extension of $G$. In both cases, let $H' = H \setminus \{x \to y\}$. We will prove that $H'$ is an extension of $G'$:

– First, it is clear that $G'$ and $H'$ have the same skeleton.

– Secondly, if $u \to v \in G'$ (note that $u \to v \neq x \to y$), then $u \to v \in G$. As $H$ is an extension of $G$, then $u \to v \in H$, and therefore $u \to v \in H'$. So, all the arcs in $G'$ are also arcs in $H'$. Moreover, every h-h pattern in $G'$ is also an h-h pattern in $H'$.

– Thirdly, if $u \to v \leftarrow w$ is an h-h pattern in $H'$ (and we know that $u \to v \leftarrow w \neq x \to y \leftarrow w$), then $u \to v \leftarrow w \in H$. As $H$ is an extension of $G$, then $u \to v \leftarrow w \in G$. Therefore, $u \to v \leftarrow w \in G'$ (the removal of the arc $x \to y$ cannot destroy any h-h pattern where $x \to y$ is not involved). So, $G'$ and $H'$ have the same h-h patterns.

In this way, $H'$ is an extension of $G'$. Moreover, we can see that $\forall u \neq y$ $Pa_{H'}(u) = Pa_H(u)$ and $Pa_{H'}(y) = Pa_H(y) \setminus \{x\}$.

(2) The scores of $G$ and $G'$ are the same as the scores of $H$ and $H'$ respectively, since $g$ is score equivalent. Moreover, as $g$ is decomposable, we can write

$$
\begin{aligned}
g(G' : D) = g(H' : D) &= \sum_u g_D(u, Pa_{H'}(u)) = \sum_{u \neq y} g_D(u, Pa_{H'}(u)) + g_D(y, Pa_{H'}(y)) = \\
&\sum_{u \neq y} g_D(u, Pa_H(u)) + g_D(y, Pa_H(y)) - g_D(y, Pa_H(y)) + g_D(y, Pa_{H'}(y)) = \\
&\sum_u g_D(u, Pa_H(u)) - g_D(y, Pa_H(y)) + g_D(y, Pa_{H'}(y)) = \\
&g(H : D) - g_D(y, Pa_H(y)) + g_D(y, Pa_{H'}(y)) = \\
&g(G : D) - g_D(y, Pa_H(y)) + g_D(y, Pa_{H'}(y))
\end{aligned}
\tag{4}
$$

Let us now consider the five different cases:

(a) In this case, we know from Table 2 that $Pa_G(y) = \emptyset$. Moreover, $Pa_{G'}(y) = \emptyset$ (because we are inserting a link) and $Pa_{H'}(y) \neq \emptyset$ (because $H'$ is an extension of $G'$ that contains the arc $x \to y$). Then, from Proposition 5 we obtain $|Pa_{H'}(y)| = 1$, i.e. $Pa_{H'}(y) = \{x\}$. Moreover, $Pa_H(y) = Pa_{H'}(y) \setminus \{x\} = \emptyset$. So, Eq. (4) becomes

$$
g(G' : D) = g(G : D) - g_D(y, \emptyset) + g_D(y, \{x\})
$$

(b) From Table 2 we get $Pa_G(y) \neq \emptyset$ or $Pa_G(x) \neq \emptyset$.

If $Pa_G(y) \neq \emptyset$, from Proposition 5 we obtain $Pa_H(y) = Pa_G(y)$. Moreover, $Pa_{H'}(y) = Pa_H(y) \cup \{x\} = Pa_G(y) \cup \{x\}$.

If $Pa_G(y) = \emptyset$ then $Pa_{G'}(y) = \{x\}$ (because we are adding the arc $x \to y$). From Proposition 5 we obtain $Pa_{H'}(y) = Pa_{G'}(y) = \{x\} = Pa_G(y) \cup \{x\}$. Moreover, $Pa_H(y) = Pa_{H'}(y) \setminus \{x\} = \emptyset = Pa_G(y)$.

In either case, Eq. (4) becomes

$$g(G' : D) = g(G : D) - g_D(y, Pa_G(y)) + g_D(y, Pa_G(y) \cup \{x\})$$

(c) In this case, $Pa_G(y) = \emptyset$ and $Pa_{G'}(y) = \{x, z\}$. From Proposition 5 we obtain $Pa_H(y) = \{x, z\}$. Moreover, $Pa_H(y) = Pa_{H'}(y) \setminus \{x\} = \{z\}$. Then, Eq. (4) becomes

$$g(G' : D) = g(G : D) - g_D(y, \{z\}) + g_D(y, \{x, z\})$$

(d) As $Pa_G(y) = \emptyset$ and $H$ is an extension of $G$ containing the arc $x \rightarrow y$, from Proposition 5 we get $Pa_H(y) = \{x\}$. Moreover, $Pa_{H'}(y) = Pa_H(y) \setminus \{x\} = \emptyset$. In this case Eq. (4) becomes

$$g(G' : D) = g(G : D) - g_D(y, \{x\}) + g_D(y, \emptyset)$$

(e) In this case, as $Pa_G(y) \neq \emptyset$, Proposition 5 asserts that $Pa_H(y) = Pa_G(y)$. Moreover, $Pa_{H'}(y) = Pa_H(y) \setminus \{x\} = Pa_G(y) \setminus \{x\}$. Therefore, Eq. (4) becomes

$$g(G' : D) = g(G : D) - g_D(y, Pa_G(y)) + g_D(y, Pa_G(y) \setminus \{x\}) \quad \square$$

## 5.1 Comparison with Other Approaches

As we have already mentioned, there are several works devoted to learning Bayesian networks, within the score+search approach, which use the space of completed PDAGs to carry out the search process. There is a slight difference between the operators considered in the different works: the addition and deletion of edges is considered by Madigan et al. (1996), within a Markov Chain Monte Carlo process, which also performs Monte Carlo sampling from the space of the orderings of the variables compatible with the current CPDAG. Edge addition and deletion is also used by Spirtes and Meek (1995), but within a greedy process that first grows the structure by adding edges and then thins it by deleting edges. Additional operators are considered by Chickering (1996), including arc reversal and creation of v-structures.

All these methods move through the space of completed PDAGs in the following way: given the current CPDAG $G$, after selecting an operator, applying it to $G$ and obtaining a neighboring PDAG $G'$, they generate a *consistent extension* $H'$ of $G'$ (a DAG belonging to the equivalence class represented by the PDAG), if one exists. If this is the case (otherwise $G'$ is not a valid configuration), then $G'$ is evaluated by computing the score of $H'$, $g(H' : D)$. The completed PDAG representation of $G'$ is then recovered from its consistent extension $H'$.

The process of checking the existence of a consistent extension and generating it is carried out with a procedure called *PDAG-to-DAG* (Dor & Tarsi, 1992), which runs in time $O(n \cdot e)$ in the worst case, where $e$ denotes the number of edges in the PDAG. Another procedure, called *DAG-to-PDAG*, is invoked in order to obtain the completed PDAG representation of the new valid configuration. There are different implementations of DAG-to-PDAG (Andersson et al., 1997; Chickering, 1995; Meek, 1995; Pearl & Verma, 1990). For example, the time complexity of the algorithm proposed by Chickering (1995) is $O(e)$ on the average and $O(n \cdot e)$ in the worst case.

Our search method does not need to use any of these two procedures: in order to check the validity of a neighboring configuration of an RPDAG $G$, it is only necessary, in some cases, to perform a test to detect either an undirected path or a partially directed path between two nodes in $G$ (implemented by procedures $UC()$ and $DC()$ in Section 4.3). On the other hand, once the search process has explored the neighborhood of $G$ and determined the best neighboring configuration $G'$, $G'$ is not always an RPDAG, and we must generate its RPDAG representation. This generation procedure is also very simple: it consists in firing, starting from a single node $y$, a cascaded process that either directs links away from $y$ or undirects arcs (implemented by procedures $Complete()$ and $Undo()$ in Section 4.3). Note that all these procedures used by our search method are less time-consuming than PDAG-to-DAG and DAG-to-PDAG.

More importantly, our search method can take advantage of the decomposability of many scoring functions, and each RPDAG (except the initial one) can be evaluated by computing only two local scores. However, the methods based on completed PDAGs need to recompute all the local scores, although the algorithm proposed by Muntenau and Cau (2000), which operates on completed PDAGs and uses three insertion operators (for arcs, links and v-structures) is also able to score any neighboring configuration using two local scores; however, the validity conditions of some of these operators are not correct.

Finally, Chickering (2002)[7] describes an algorithm that searches in the space of completed PDAGs and is also able to evaluate configurations by computing only (up to four) local scores. It uses six operators, link and arc addition, link and arc deletion, creation of v-structures by directing two already existing links, and reversal of arcs. All the operators can be evaluated using two local scores, except reversal and creation of v-structures, that require four local scores. The validity conditions of the operators are established essentially in terms of two conditions: (1) the absence of semi-directed or undirected paths between two nodes that do not pass through certain set of nodes, $S$, and (2) the fact that a certain set of nodes forms a clique. Link insertion and creation of v-structures need the first type of condition, link and arc deletion need the second one, whereas arc insertion and arc reversal require both conditions. The "path" validity conditions take time $O(|S| + e)$ in the worst case, and the "clique" conditions take time $O(|S|^2)$, also in the worst case. This algorithm also requires the $PDAG\text{-}to\text{-}DAG$ and $DAG\text{-}to\text{-}PDAG$ procedures to be used.

So, although the validity conditions of the operators in Chickering's algorithm and their postprocessing are somewhat more complex than ours, the advantage is that this algorithm does not have any duplicate representations of the equivalence classes. Whether the computational cost of moves in the CPDAG space can compensate for the larger number of RPDAGs (and the larger number of local scores to be computed) is a matter of empirical evaluation, that will possibly depend on the "sparseness" of the specific domain problem considered.

# 6. Experimental Results

In this section we shall describe the experiments carried out with our algorithm, the obtained results, and a comparative study with other algorithms for learning Bayesian networks. We have selected nine different problems to test our algorithm, all of which only contain discrete variables: Alarm (Figure 18), Insurance (Figure 19), Hailfinder (Figure 20), Breast-Cancer, crx, Flare2, House-Votes, Mushroom, and Nursery.

---

7. This work appeared after the original submission of this paper.

The Alarm network displays the relevant variables and relationships for the Alarm Monitoring System (Beinlich et al., 1989), a diagnostic application for patient monitoring. This network, which contains 37 variables and 46 arcs, has been considered as a benchmark for evaluating Bayesian network learning algorithms. The input data commonly used are subsets of the Alarm database built by Herskovits (1991), which contains 20000 cases that were stochastically generated using the Alarm network. In our experiments, we have used three databases of different sizes (the first $k$ cases in the Alarm database, for $k = 3000, 5000$ and 10000).

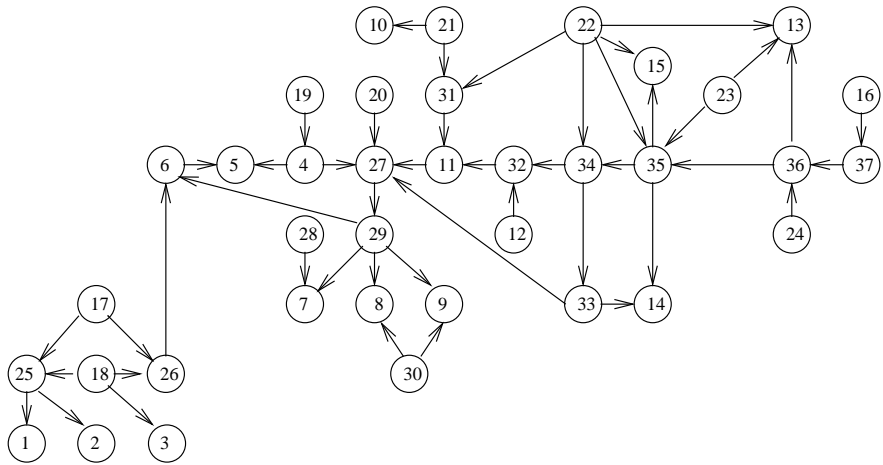

Figure 18: The Alarm network

Insurance (Binder et al., 1997) is a network for evaluating car insurance risks. The Insurance network contains 27 variables and 52 arcs. In our experiments, we have used five databases containing 10000 cases, generated from the Insurance Bayesian network.

Hailfinder (Abramson et al., 1996) is a normative system that forecasts severe summer hail in northeastern Colorado. The Hailfinder network contains 56 variables and 66 arcs. In this case, we have also used five databases with 10000 cases generated from the Hailfinder network.

Breast-Cancer, crx, Flare2, House-Votes, Mushroom, and Nursery are databases available from the UCI Machine Learning Repository. Breast-Cancer contains 10 variables (9 attributes, two of which have missing values, and a binary class variable) and 286 instances. The crx database concerns credit card applications. It has 490 cases and 16 variables (15 attributes and a class variable), and seven variables have missing values. Moreover, six of the variables in the crx database are continuous and were discretized using the MLC++ system (Kohavi, John, Long, Manley & Pfleger, 1994). Flare2 uses 13 variables (10 attributes and 3 class variables, one for the number of times a certain type of solar flare occured in a 24-hour period) and contains 1066 instances, without missing values. House-Votes stores the votes for each of the U.S. House of Representatives Congressmen on 16 key votes; it has 17 variables and 435 records and all the variables except two have missing values. Mushroom contains 8124 cases corresponding to species of gilled mushrooms in the Agaricus and Lepiota Family; there are 23 variables (a class variable, stating whether the mushroom is edible or poisonous, and 22 attribute variables) and only one variable has missing values.

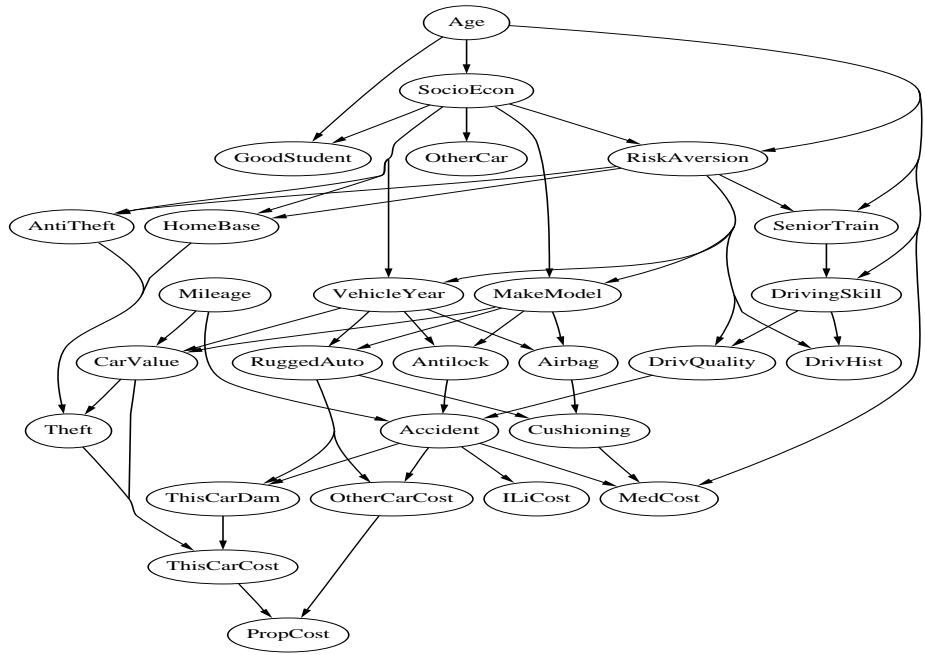

Figure 19: The Insurance network

Nursery contains data relative to the evaluation of applications for nursery schools, and has 9 variables and 12960 cases, without missing values. In all of the cases, missing values are not discarded but treated as a distinct state.

In the first series of experiments, we aim to compare the behavior of our RPDAG-based local search method (RPDAG) with the classical local search in the space of DAGs (DAG). The scoring function selected is BDeu (Heckerman et al., 1995) (which is score equivalent and decomposable), with the parameter representing the equivalent sample size set to 1 and a uniform structure prior. The starting point of the search is the empty graph in both cases.

We have collected the following information about the experiments:

**BDeu.-** The BDeu score (log version) of the learned network.

**Edg.-** The number of edges included in the learned network.

**H.-** The Hamming distance, H=A+D+I, i.e. the number of different edges, added (A), deleted (D), or wrongly oriented (without taking into account the differences between equivalent structures) (I), in the learned network with respect to the gold standard network (the original model). This measure is only computed for the three test domains where a gold standard exists.

**Iter.-** The number of iterations carried out by the algorithm to reach the best network, i.e. the number of operators used to transform the initial graph into a local optimum.

**Ind.-** The number of individuals (either DAGs or RPDAGs) evaluated by the algorithm.

Figure 20: The Hailfinder network

**EstEv.-** The number of *different* statistics evaluated during the execution of the algorithm. This is a useful value to measure the efficiency of the algorithms, because most of the running time of a scoring-based learning algorithm is spent in the evaluation of statistics from the database.

**TEst.-** The total number of statistics used by the algorithm. Note that this number can be considerably greater than EstEv. By using hashing techniques we can store and efficiently retrieve any previously calculated statistics. It is not therefore necessary to recompute them by accessing the database, thus gaining in efficiency.

**NVars.-** The average number of variables that intervene in the different statistics (i.e. the values $N_{y,Pa_H(y)}$ in Eq. (3)) computed. This value is also important because the time required to compute a statistic increases exponentially with the number of variables involved.

**Time.-** The time, measured in *seconds*, employed by the algorithm to learn the network. Our implementation is written in the JAVA programming language and runs under Linux. This value is only a rough measure of the efficiency of the algorithms, because there are many circumstances that may influence the running time (external loading in a networked computer, caching or any other aspect of the computer architecture, memory paging, use of virtual memory, threading, different code, etc.). Nevertheless, we have tried to ensure that the two algorithms run under the same conditions as far as possible, and the two implementations share most of the code. In fact, the two algorithms have been integrated into the Elvira package (available at `http://www.leo.ugr.es/~elvira`).

For the Insurance and Hailfinder domains, the reported results are the average values across the five databases considered. The results of our experiments for synthetic data, i.e. Alarm, Insurance and Hailfinder, are displayed in Tables 3, 4 and 5, respectively, where we also show the BDeu values for the true ($T_D$) and the empty ($\emptyset_D$) networks (with parameters re-trained from the corresponding database $D$), which may serve as a kind of scale. The results obtained for real data are displayed in Table 6.

As we consider there to be a clear difference between the results obtained for the synthetic and the UCI domains, we shall discuss them separately.

- For the synthetic domains (Tables 3, 4 and 5):

    – Our RPDAG-based algorithm outperforms the DAG-based one with respect to the value of the scoring function used to guide the search: we always obtain better results on the five databases considered. Note that we are using a logarithmic version of the scoring function, so that the differences are much greater in a non-logarithmic scale. These results support the idea that RPDAGs are able to find new and better local maxima within the score+search approach for learning Bayesian networks in this type of highly structured domains.

    – Our search method is also preferable from the point of view of the Hamming distances, which are always considerably lower than the ones obtained by using the DAG space.

    – Moreover, our search method is generally more efficient: it carries out fewer iterations (on the five cases), evaluates fewer individuals (on four cases), computes

| | BDeu | Edg | H | A | D | I | Iter | Ind | EstEv | TEst | NVar | Time |
|---|---|---|---|---|---|---|---|---|---|---|---|---|
| Alarm-3000 | | | | | | | | | | | | |
| RPDAG | -33101 | 46 | 2 | 1 | 1 | 0 | 49 | 63124 | 3304 | 123679 | 2.98 | 111 |
| DAG | -33109 | 47 | 7 | 3 | 2 | 2 | 58 | 72600 | 3300 | 145441 | 2.88 | 117 |
| $T_{\text{Alarm3}}$ | -33114 | 46 | | | | | | | | | | |
| $\emptyset_{\text{Alarm3}}$ | -59890 | 0 | | | | | | | | | | |
| Alarm-5000 | | | | | | | | | | | | |
| RPDAG | -54761 | 46 | 2 | 1 | 1 | 0 | 49 | 62869 | 3326 | 123187 | 2.97 | 179 |
| DAG | -54956 | 54 | 16 | 9 | 1 | 6 | 60 | 76212 | 3391 | 152663 | 2.93 | 194 |
| $T_{\text{Alarm5}}$ | -54774 | 46 | | | | | | | | | | |
| $\emptyset_{\text{Alarm5}}$ | -99983 | 0 | | | | | | | | | | |
| Alarm-10000 | | | | | | | | | | | | |
| RPDAG | -108432 | 45 | 1 | 0 | 1 | 0 | 48 | 61190 | 3264 | 120049 | 2.97 | 346 |
| DAG | -108868 | 52 | 13 | 7 | 1 | 5 | 60 | 75504 | 3449 | 151251 | 2.94 | 380 |
| $T_{\text{Alarm10}}$ | -108452 | 46 | | | | | | | | | | |
| $\emptyset_{\text{Alarm10}}$ | -199920 | 0 | | | | | | | | | | |

Table 3: Results for the Alarm databases

| | BDeu | Edg | H | A | D | I | Iter | Ind | EstEv | TEst | NVar | Time |
|---|---|---|---|---|---|---|---|---|---|---|---|---|
| RPDAG | -133071 | 45 | 18 | 4 | 10 | 4 | 48 | 36790 | 1990 | 69965 | 2.95 | 202 |
| DAG | -133205 | 49 | 25 | 7 | 10 | 8 | 58 | 36178 | 2042 | 72566 | 3.03 | 214 |
| $T_{\text{Insur}}$ | -133040 | 52 | | | | | | | | | | |
| $\emptyset_{\text{Insur}}$ | -215278 | 0 | | | | | | | | | | |

Table 4: Average results for the Insurance domain across 5 databases of size 10000

fewer different statistics from the databases (on three cases), uses fewer statistics (on the five cases), and runs faster (on four cases). On the contrary, the average number of variables involved in the statistics is slightly greater (on four cases).

- For the UCI domains (Table 6):

    - The results in this case are not as conclusive about the advantages of the RPDAG-based method with respect to the DAG-based one in terms of effectiveness: both

| | BDeu | Edg | H | A | D | I | Iter | Ind | EstEv | TEst | NVar | Time |
|---|---|---|---|---|---|---|---|---|---|---|---|---|
| RPDAG | -497872 | 67 | 24 | 12 | 10 | 2 | 68 | 235374 | 7490 | 459436 | 2.92 | 847 |
| DAG | -498395 | 75 | 45 | 21 | 13 | 11 | 81 | 240839 | 7313 | 482016 | 2.81 | 828 |
| $T_{\text{Hail}}$ | -503230 | 66 | | | | | | | | | | |
| $\emptyset_{\text{Hail}}$ | -697826 | 0 | | | | | | | | | | |

Table 5: Average results for the Hailfinder domain across 5 databases of size 10000

|  | BDeu | Edg | Iter | Ind | EstEv | TEst | NVar | Time |
|---|---|---|---|---|---|---|---|---|
| Breast-Cancer | | | | | | | | |
| RPDAG | -2848 | 6 | 7 | 619 | 151 | 1232 | 2.26 | 0.673 |
| DAG | -2848 | 6 | 7 | 619 | 148 | 1284 | 2.26 | 0.686 |
| crx | | | | | | | | |
| RPDAG | -5361 | 19 | 20 | 5318 | 545 | 10020 | 2.75 | 3.03 |
| DAG | -5372 | 19 | 20 | 4559 | 510 | 9208 | 2.61 | 2.85 |
| Flare2 | | | | | | | | |
| RPDAG | -6728 | 15 | 16 | 2637 | 329 | 4887 | 2.71 | 3.66 |
| DAG | -6733 | 13 | 14 | 2012 | 310 | 4093 | 2.45 | 3.23 |
| House-Votes | | | | | | | | |
| RPDAG | -4629 | 22 | 23 | 6370 | 591 | 11883 | 2.80 | 3.10 |
| DAG | -4643 | 23 | 24 | 6094 | 621 | 12289 | 2.66 | 3.39 |
| Mushroom | | | | | | | | |
| RPDAG | -77239 | 92 | 97 | 43121 | 2131 | 78459 | 3.99 | 432 |
| DAG | -77208 | 87 | 103 | 39944 | 2173 | 80175 | 3.96 | 449 |
| Nursery | | | | | | | | |
| RPDAG | -125717 | 8 | 9 | 415 | 115 | 803 | 2.75 | 11.91 |
| DAG | -125717 | 8 | 9 | 611 | 133 | 1269 | 2.59 | 13.50 |

Table 6: Results for the UCI databases

algorithms reach the same solution on two cases from six, RPDAG is better than DAG on three cases, and DAG is better on one case.

– With respect to the efficiency of the two algorithms, the situation is similar: neither algorithm clearly outperforms the other with respect to any of the five efficiency measures considered.

In a second series of experiments, we aim to test the behavior of the search in the RPDAG space when used in combination with a search heuristic which is more powerful than a simple greedy search. The heuristic selected is *Tabu Search* (Glover, 1989; Bouckaert, 1995), which tries to escape from a local maximum by selecting a solution that minimally decreases the value of the scoring function; immediate re-selection of the local maximum just visited is prevented by maintaining a list of solutions that are forbidden, the so-called *tabu list* (although for practical reasons the tabu list stores forbidden operators not solutions, and consequently, solutions which have not been visited previously may also become forbidden).

We have implemented two simple versions of tabu search: TS-RPDAG and TS-DAG, which explore the RPDAG and DAG spaces, respectively, using the same operators as their respective greedy versions. The parameters used by these algorithms are the length *tll* of the tabu list and the number *tsit* of iterations required to stop the search process. In our experiments, these values have been fixed as follows: $tll = n$ and $tsit = n(n-1)$, $n$ being the number of variables in the domain. The scoring function and the initial graph are the same as in previous experiments, as well as the collected performance measures, with one exception: as the number of iterations is now fixed (Iter=$tsit$), we compute the iteration where the best graph was found (BIter) instead. The results of these experiments are displayed in Tables 7, 8, 9 and 10.

| | BDeu | Edg | H | A | D | I | BIter | Ind | EstEv | TEst | NVar | Time |
|---|---|---|---|---|---|---|---|---|---|---|---|---|
| | | | | | Alarm-3000 | | | | | | | |
| TS-RPDAG | -33101 | 46 | 2 | 1 | 1 | 0 | 48 | 1779747 | 9596 | 3044392 | 3.63 | 286 |
| TS-DAG | -33115 | 51 | 11 | 7 | 2 | 2 | 129 | 1320696 | 8510 | 2645091 | 3.59 | 391 |
| | | | | | Alarm-5000 | | | | | | | |
| TS-RPDAG | -54761 | 46 | 2 | 1 | 1 | 0 | 48 | 1779579 | 10471 | 3031966 | 3.61 | 421 |
| TS-DAG | -54762 | 47 | 3 | 2 | 1 | 0 | 720 | 1384990 | 11113 | 2773643 | 3.58 | 541 |
| | | | | | Alarm-10000 | | | | | | | |
| TS-RPDAG | -108432 | 45 | 1 | 0 | 1 | 0 | 47 | 1764670 | 10671 | 3020165 | 3.66 | 735 |
| TS-DAG | -108442 | 50 | 6 | 5 | 1 | 0 | 284 | 1385065 | 11014 | 2773795 | 3.60 | 862 |

Table 7: Results for the Alarm databases using Tabu Search

| | BDeu | Edg | H | A | D | I | BIter | Ind | EstEv | TEst | NVar | Time |
|---|---|---|---|---|---|---|---|---|---|---|---|---|
| TS-RPDAG | -133070 | 45 | 18 | 4 | 10 | 4 | 58 | 458973 | 2823 | 751551 | 3.44 | 182 |
| TS-DAG | -132788 | 47 | 18 | 5 | 10 | 3 | 415 | 352125 | 5345 | 706225 | 4.16 | 428 |

Table 8: Average results for the Insurance domain using Tabu Search

| | BDeu | Edg | H | A | D | I | BIter | Ind | EstEv | TEst | NVar | Time |
|---|---|---|---|---|---|---|---|---|---|---|---|---|
| TS-RPDAG | -497872 | 67 | 24 | 12 | 10 | 2 | 67 | 9189526 | 19918 | 1.5223387E7 | 4.07 | 3650 |
| TS-DAG | -498073 | 70 | 35 | 17 | 13 | 5 | 1631 | 7512114 | 22184 | 1.5031642E7 | 4.07 | 4513 |

Table 9: Average results for the Hailfinder domain using Tabu Search

For the synthetic domains, in all the cases, except in one of the insurance databases, the results obtained by TS-RPDAG and RPDAG are the same. This phenomenon also appears for the UCI databases, where only in two databases does TS-RPDAG improve the results of RPDAG. Therefore, the Tabu Search does not contribute significantly to improving the greedy search in the RPDAG space (at least using the selected values for the parameters *tll* and *tsit*). This is in contrast with the situation in the DAG space, where TS-DAG improves the results obtained by DAG, with the exception of two UCI databases (equal results) and Alarm-3000 (where DAG performs better than TS-DAG).

With respect to the comparison between TS-RPDAG and TS-DAG, we still consider TS-RPDAG to be preferable to TS-DAG on the synthetic domains, although in this case TS-DAG performs better on the insurance domain. For the UCI databases, the two algorithms perform similarly: each algorithm is better than the other on two domains, and both algorithms perform equally on the remaining two domains. TS-RPDAG is somewhat more efficient than TS-DAG with respect to running time.

|         | BDeu    | Edg | BIter | Ind    | EstEv | TEst   | NVar | Time  |
|---------|---------|-----|-------|--------|-------|--------|------|-------|
| Breast-Cancer | | | | | | | | |
| TS-RPDAG | -2848  | 6   | 6     | 8698   | 345   | 14209  | 3.03 | 1.96  |
| TS-DAG   | -2848  | 6   | 6     | 6806   | 316   | 13892  | 2.87 | 1.98  |
| crx     | | | | | | | | |
| TS-RPDAG | -5361  | 19  | 19    | 61175  | 908   | 98574  | 3.20 | 9.26  |
| TS-DAG   | -5362  | 20  | 29    | 44507  | 1176  | 89714  | 3.17 | 12.23 |
| Flare2  | | | | | | | | |
| TS-RPDAG | -6728  | 15  | 15    | 23363  | 616   | 37190  | 3.47 | 9.70  |
| TS-DAG   | -6726  | 15  | 129   | 18098  | 681   | 36665  | 3.27 | 10.60 |
| House-Votes | | | | | | | | |
| TS-RPDAG | -4622  | 24  | 180   | 73561  | 1144  | 121206 | 3.32 | 11.14 |
| TS-DAG   | -4619  | 23  | 252   | 56570  | 1364  | 113905 | 3.29 | 15.30 |
| Mushroom | | | | | | | | |
| TS-RPDAG | -77002 | 99  | 495   | 209556 | 4021  | 350625 | 4.83 | 883   |
| TS-DAG   | -77073 | 90  | 450   | 157280 | 3455  | 315975 | 4.57 | 1725  |
| Nursery | | | | | | | | |
| TS-RPDAG | -125717 | 8  | 8     | 4352   | 251   | 6525   | 3.09 | 28.05 |
| TS-DAG   | -125717 | 8  | 8     | 3898   | 237   | 7991   | 2.95 | 26.20 |

Table 10: Results for the UCI databases using Tabu Search

We have carried out a third series of experiments to compare our learning algorithm based on RPDAGs with other algorithms for learning Bayesian networks. In this case, the comparison is only intended to measure the quality of the learned network. In addition to the DAG-based local and tabu search previously considered, we have also used the following algorithms:

- PC (Spirtes et al., 1993), an algorithm based on independence tests. We used an independence test based on the measure of conditional mutual information (Kullback, 1968), with a fixed confidence level equal to 0.99.

- The K2 search method (Cooper & Herskovits, 1992), in combination with the BDeu scoring function (K2). Note that K2 needs an ordering of the variables as the input. We used an ordering consistent with the topology of the corresponding networks.

- Another algorithm, BN Power Constructor (BNPC), that uses independence tests (Cheng et al., 1997; Cheng, Bell & Liu, 1998).

The two independence-based algorithms, PC and BNPC, operate on the space of equivalence classes, whereas K2 explores the space of DAGs which are compatible with a given ordering. We have included the algorithm K2 in the comparison, using a correct ordering and the same scoring function as the RPDAG and DAG-based search methods, in order to test whether our method can outperform the results obtained with a more informed algorithm. The results for the algorithms PC and K2 have been obtained using our own implementations (which are also included in the Elvira software). For BNPC, we used the software package available at http://www.cs.ualberta.ca/~jcheng/bnsoft.htm.

The test domains included in these experiments are Alarm, Insurance, and Hailfinder. In addition to the BDeu values, the number of edges in the learned networks and the Hamming distances, we have collected two additional performance measures:

**BIC.-** The value of the BIC (Bayesian Information Criterion) scoring function (Schwarz, 1978) for the learned network. This value measures the quality of the network using maximum likelihood and a penalty term. Note that BIC is also score-equivalent and decomposable.

**KL.-** The Kullback-Leibler distance (cross-entropy) (Kullback, 1968) between the probability distribution, $P$, associated to the database (the empirical frequency distribution) and the probability distribution associated to the learned network, $P_G$. Notice that this measure is actually the same as the log probability of the data. We have in fact calculated a decreasing monotonic linear transformation of the Kullback-Leibler distance, because this one has exponential complexity and the transformation can be computed very efficiently: If $P_G$ is the joint probability distribution associated to a network $G$, then the KL distance can be written in the following way (de Campos, 1998; Lam & Bacchus, 1994):

$$KL(P, P_G) = -H_P(\mathcal{U}) + \sum_{x \in \mathcal{U}} H_P(x) - \sum_{x \in \mathcal{U}, Pa_G(x) \neq \emptyset} MI_P(x, Pa_G(x)) \qquad (5)$$

where $H_P(Z)$ denotes Shannon entropy with respect to the distribution $P$ for the subset of variables $Z$ and $MI_P(x, Pa_G(x))$ is the measure of mutual information between the two sets of variables $\{x\}$ and $Pa_G(x)$. As the first two terms of the expression above do not depend on the graph $G$, our transformation consists in calculating only the third term in equation (5). So, the interpretation of our transformation of the Kullback-Leibler distance is: the higher this value is, the better the network fits the data. However, this measure should be handled with caution, since a high KL value may also indicate overfitting (a network with many edges will probably have a high KL value).

Although for those algorithms whose goal is to optimize the Bayesian score, BDeu is really the metric that should be used to evaluate them, we have also computed BIC and KL because two of the algorithms considered use independence tests instead of a scoring function.

The results of these experiments are displayed in Table 11. The best value for each performance measure and each database is written in bold, and the second best value in italics. These results indicate that our search method in the RPDAG space, in combination with the BDeu scoring function, is competitive with respect to other algorithms: only the TS-DAG algorithm, which uses a more powerful (and more computationally intensive) search heuristic in the DAG space, and, to a lesser extent, the more informed K2 algorithm, perform better than RPDAG in some cases. Observe that both TS-DAG and K2 perform better than RPDAG in terms of KL on four cases from five.

|  | BDeu | BIC | KL | Edg | H | A | D | I |
|---|---|---|---|---|---|---|---|---|
| \multicolumn{9}{c}{Alarm-3000} |
| RPDAG | **-33101** | **-33930** | *9.23055* | 46 | **2** | 1 | 1 | 0 |
| DAG | *-33109* | *-33939* | 9.23026 | 47 | *7* | 3 | 2 | 2 |
| TS-DAG | -33115 | -33963 | 9.23047 | 51 | 11 | 7 | 2 | 2 |
| PC | -36346 | -36691 | 8.06475 | 37 | 10 | 0 | 9 | 1 |
| BNPC | -33422 | -35197 | 9.11910 | 43 | 7 | 2 | 5 | 0 |
| K2 | -33127 | -34351 | **9.23184** | 46 | **2** | 1 | 1 | 0 |
| \multicolumn{9}{c}{Alarm-5000} |
| RPDAG | **-54761** | **-55537** | 9.25703 | 46 | **2** | 1 | 1 | 0 |
| DAG | -54956 | -55831 | 9.25632 | 54 | 16 | 9 | 1 | 6 |
| TS-DAG | *-54762* | *-55540* | *9.25736* | 47 | *3* | 2 | 1 | 0 |
| PC | -61496 | -61822 | 7.85435 | 38 | 16 | 2 | 10 | 4 |
| BNPC | -55111 | -55804 | 9.16787 | 42 | 4 | 0 | 4 | 0 |
| K2 | -54807 | -55985 | **9.25940** | 47 | *3* | 2 | 1 | 0 |
| \multicolumn{9}{c}{Alarm-10000} |
| RPDAG | **-108432** | **-109165** | 9.27392 | 45 | **1** | 0 | 1 | 0 |
| DAG | -108868 | -110537 | **9.27809** | 52 | 13 | 7 | 1 | 5 |
| TS-DAG | *-108442* | *-109188* | 9.27439 | 50 | 6 | 5 | 1 | 0 |
| PC | -117661 | -117914 | 8.31704 | 38 | 11 | 1 | 9 | 1 |
| BNPC | -109164 | -109827 | 9.18884 | 42 | 4 | 0 | 4 | 0 |
| K2 | -108513 | -109647 | *9.27549* | 46 | *2* | 1 | 1 | 0 |
| \multicolumn{9}{c}{Insurance} |
| RPDAG | -133071 | -134495 | 8.38502 | 45 | *18* | 4 | 10 | 4 |
| DAG | -133205 | -135037 | 8.39790 | 49 | 25 | 7 | 10 | 8 |
| TS-DAG | *-132788* | *-134414* | *8.41467* | 47 | *18* | 5 | 10 | 3 |
| PC | -139101 | -141214 | 7.75574 | 33 | 23 | 0 | 20 | 3 |
| BNPC | -134726 | -135832 | 8.21606 | 37 | 26 | 3 | 18 | 5 |
| K2 | **-132615** | **-134095** | **8.42471** | 44 | **10** | 1 | 9 | 0 |
| \multicolumn{9}{c}{Hailfinder} |
| RPDAG | **-497872** | -531138 | 20.53164 | 67 | *24* | 12 | 10 | 2 |
| DAG | -498395 | -531608 | 20.48503 | 75 | 45 | 21 | 13 | 11 |
| TS-DAG | *-498073* | *-516100* | *20.59372* | 70 | 35 | 17 | 13 | 5 |
| PC | -591507 | -588638 | 12.65981 | 49 | 50 | 16 | 33 | 1 |
| BNPC | -503440 | **-505160** | **20.61581** | 64 | 28 | 12 | 15 | 1 |
| K2 | -498149 | -531373 | 20.51822 | 67 | **23** | 12 | 11 | 0 |

Table 11: Performance measures for different learning algorithms

The fourth series of experiments attempts to evaluate the behavior of the same algorithms on a dataset which is different from the training set used to learn the network. In order to do so, we have computed the BDeu, BIC, and KL values of the network structure learned using a database, with respect to a different database: for the Alarm domain, the training set is the Alarm-3000 database used previously, and the test set is formed by the 3000 next cases in the Alarm database; for both Insurance and Hailfinder, we selected one of the five databases that we have been using as the training set and another of these databases as the test set. The results are shown in Table 12. We can observe that they are analogous to the results obtained in Table 11, where the same databases were used for training and testing. However, in this case RPDAG also performs better than TS-DAG and K2 in terms of KL. Therefore, the good behavior of our algorithm cannot be attributed to overfitting.

| | BDeu | BIC | KL |
|---|---|---|---|
| Alarm-3000 | | | |
| RPDAG | **-32920** | **-33750** | _9.35839_ |
| DAG | _-32938_ | _-33769_ | 9.35488 |
| TS-DAG | -32947 | -33793 | 9.35465 |
| PC | -36286 | -36632 | 8.15227 |
| BNPC | -33309 | -35051 | 9.23576 |
| K2 | -32951 | -34165 | **9.36168** |
| Insurance | | | |
| RPDAG | -132975 | -134471 | **8.44891** |
| DAG | -133004 | -134952 | _8.44745_ |
| TS-DAG | **-132810** | **-134281** | 8.44538 |
| PC | -140186 | -143011 | 7.70182 |
| BNPC | -135029 | -136125 | 8.23063 |
| K2 | _-132826_ | _-134309_ | 8.44671 |
| Hailfinder | | | |
| RPDAG | **-497869** | -531165 | _20.58267_ |
| DAG | -498585 | -531913 | 20.49730 |
| TS-DAG | _-497983_ | **-505592** | 20.55420 |
| PC | -587302 | -584939 | 12.68256 |
| BNPC | -506680 | _-508462_ | **20.71069** |
| K2 | -498118 | -531356 | 20.57876 |

Table 12: Performance measures for the learning algorithms using a different test set

Finally, we have carried out another series of experiments, which aim to compare our RPDAG algorithm with the algorithm proposed by Chickering (2002), that searches in the CPDAG space. In this case, we have selected the House-Votes and Mushroom domains (which were two of the datasets used by Chickering). In order to approximate our experimental conditions to those described in Chickering's work, we used the BDeu scoring function with a prior equivalent sample size of ten, and a structure prior of $0.001^f$, where $f$ is the number of free parameters in the DAG; moreover, we used five random subsets of the original databases, each containing approximately 70% of the total data (304 cases for House-Votes and 5686 for Mushroom). Table 13 displays the average values across the five datasets of the relative improvement of the per-case score obtained by RPDAG to the per-case score of DAG, as well as the ratio of the time spent by DAG to the time spent by RPDAG. We also show in Table 13 the corresponding values obtained by Chickering (using only one dataset) for the comparison between his CPDAG algorithm and DAG.

We may observe that the behavior of RPDAG and CPDAG is somewhat different: although both algorithms are more efficient than DAG, it seems to us that CPDAG runs faster than RPDAG. With respect to effectiveness, both RPDAG and CPDAG obtain exactly the same solution as DAG in the House-Votes domain (no relative improvement); however, in the other domain, RPDAG outperforms DAG (on the five datasets considered) whereas CPDAG performs worse than DAG. In any case, the differences are small (they could be a result of differences in the experimental setup) and a much more systematic experimentation with these algorithms would be necessary in order to establish general conclusions about their comparative behavior.

| | RPDAG versus DAG | | CPDAG versus DAG | |
|---|---|---|---|---|
| Dataset | Relative Improv. | Time Ratio | Relative Improv. | Time Ratio |
| House-Votes | 0.0000 | 1.041 | 0.0000 | 1.27 |
| Mushroom | 0.0158 | 1.005 | -0.0382 | 2.81 |

Table 13: Comparison with Chickering's work on completed PDAGs

## 7. Concluding Remarks

We have developed a new local search algorithm, within the score+search approach for learning Bayesian network structures from databases. The main feature of our algorithm is that it does not search in the space of DAGs, but uses a new form of representation, restricted PDAGs, that allows us to search efficiently in a space similar to the space of equivalence classes of DAGs. For the common situation in which a decomposable scoring function is used, the set of operators that define the neighborhood structure of our search space can be scored locally (as it happens in the space of DAGs), i.e. we can evaluate any neighboring restricted PDAG by computing at most two local scores. In this way, we maintain the computational efficiency that the space of DAGs offers and, at the same time, we explore a more reduced search space, with a smoother landscape, which avoids some early decisions on edge directions. These characteristics may help to direct the search process towards better network structures.

The experimental results show that our search method based on restricted PDAGs can efficiently and accurately recover complex Bayesian network structures from data, and can compete with several state of the art Bayesian network learning algorithms, although it does not significantly improve them. Our experiments in Section 6, as well as those conducted by Chickering (2002), seem to point out that search algorithms based on PDAGs can obtain slightly better results, with respect to both effectiveness and efficiency, than search methods based on DAGs, especially for highly structured models (i.e., models that can be (almost) perfectly represented by a DAG). We believe that PDAGs can also be useful in domains which are complex (contain many variables and complicated dependence patterns) and sparse (represent many independence relationships).

For future research, we are planning to integrate the techniques developed in this paper within more powerful search methods, such as the ones considered by Blanco et al. (2003), de Campos et al. (2002) or de Campos and Puerta (2001a). Additionally, in the light of the results obtained by our method in combination with Tabu Search, it may be interesting to incorporate another operator, which could either be a classical arc reversal or some kind of specific operator to destroy h-h patterns. We also intend to work on the adaptation and application of our algorithm to real problems in the field of classification.

## Acknowledgments

This work has been supported by the Spanish Ministerio de Ciencia y Tecnología (MCYT) and the Junta de Comunidades de Castilla-La Mancha under Projects TIC2001-2973-CO5-01 and PBC-02-002, respectively. We are grateful to our colleague José M. Puerta for his invaluable help with the implementation of several algorithms. We are also grateful to the three anonymous reviewers for useful comments and suggestions.

## References

Abramson, B., Brown, J., Murphy, A., & Winkler, R. L. (1996). Hailfinder: A Bayesian system for forecasting severe weather. *International Journal of Forecasting, 12*, 57–71.

Acid, S., & de Campos, L. M. (2000). Learning right sized belief networks by means of a hybrid methodology. *Lecture Notes in Artificial Intelligence, 1910*, 309–315.

Acid, S., & de Campos, L. M. (2001). A hybrid methodology for learning belief networks: Benedict. *International Journal of Approximate Reasoning, 27*, 235–262.

Andersson, S., Madigan, D., & Perlman, M. (1997). A Characterization of Markov equivalence classes for acyclic digraphs. *Annals of Statistics, 25*, 505–541.

Beinlich, I. A., Suermondt, H. J., Chavez, R. M., & Cooper, G. F. (1989). The alarm monitoring system: A case study with two probabilistic inference techniques for belief networks. *In Proceedings of the European Conference on Artificial Intelligence in Medicine*, 247–256.

Binder, J., Koller, D., Russell, S., & Kanazawa, K. (1997). Adaptive probabilistic networks with hidden variables. *Machine Learning, 29*, 213–244.

Blanco, R., Inza, I., & Larrañaga, P. (2003). Learning Bayesian networks in the space of structures by estimation of distribution algorithms. *International Journal of Intelligent Systems, 18*, 205–220.

Bouckaert, R. R. (1993). Belief networks construction using the minimum description length principle. *Lecture Notes in Computer Science, 747*, 41–48.

Bouckaert, R. R. (1995). *Bayesian belief networks: from construction to inference*. Ph.D. thesis, University of Utrecht.

Buntine, W. (1991). Theory refinement of Bayesian networks. *In Proceedings of the Seventh Conference on Uncertainty in Artificial Intelligence*, 52–60.

Buntine, W. (1994). Operations for learning with graphical models. *Journal of Artificial Intelligence Research, 2*, 159–225.

Buntine, W. (1996). A guide to the literature on learning probabilistic networks from data. *IEEE Transactions on Knowledge and Data Engineering, 8,* 195–210.

Cheng, J., Bell, D. A., & Liu, W. (1997). An algorithm for Bayesian belief network construction from data. *In Proceedings of AI and STAT'97,* 83–90.

Cheng, J., Bell, D. A., & Liu, W. (1998). Learning Bayesian networks from data: An efficient approach based on information theory. Tech. rep., University of Alberta.

Chickering, D. M. (1995). A transformational characterization of equivalent Bayesian network structures. *In Proceedings of the Eleventh Conference on Uncertainty in Artificial Intelligence,* 87–98.

Chickering, D. M. (1996). Learning equivalence classes of Bayesian network structures. *In Proceedings of the Twelfth Conference on Uncertainty in Artificial Intelligence,* 150–157.

Chickering, D. M. (2002). Learning equivalence classes of Bayesian network structures. *Journal of Machine Learning Research, 2,* 445–498.

Chickering, D. M., Geiger, D., & Heckerman, D. (1995). Learning Bayesian networks: Search methods and experimental results. *In Preliminary Papers of the Fifth International Workshop on Artificial Intelligence and Statistics,* 112–128.

Chow, C., & Liu, C. (1968). Approximating discrete probability distributions with dependence trees. *IEEE transactions on Information Theory, 14,* 462–467.

Cooper, G. F., & Herskovits, E. (1992). A Bayesian method for the induction of probabilistic networks from data. *Machine Learning, 9,* 309–348.

Dash, D., & Druzdzel, M. (1999). A hybrid anytime algorithm for the construction of causal models from sparse data. *In Proceedings of the Fifteenth Conference on Uncertainty in Artificial Intelligence,* 142–149.

de Campos, L. M. (1998). Independency relationships and learning algorithms for singly connected networks. *Journal of Experimental and Theoretical Artificial Intelligence, 10,* 511–549.

de Campos, L. M., Fernández-Luna, J. M., Gámez, J. A., & Puerta, J. M. (2002). Ant colony optimization for learning Bayesian networks. *International Journal of Approximate Reasoning, 31,* 291–311.

de Campos, L. M., Fernández-Luna, J. M., & Puerta, J. M. (2002). Local search methods for learning Bayesian networks using a modified neighborhood in the space of dags. *Lecture Notes in Computer Science, 2527,* 182–192.

de Campos, L. M., Fernández-Luna, J. M., & Puerta, J. M. (2003). An iterated local search algorithm for learning Bayesian networks with restarts based on conditional independence tests. *International Journal of Intelligent Systems, 18,* 221–235.

de Campos, L. M., Gámez, J. A., & Puerta, J. M. (in press). Learning Bayesian networks by ant colony optimisation: Searching in two different spaces. *Mathware and Soft Computing*.

de Campos, L. M., & Huete, J. F. (1997). On the use of independence relationships for learning simplified belief networks. *International Journal of Intelligent Systems, 12*, 495–522.

de Campos, L. M., & Huete, J. F. (2000). A new approach for learning belief networks using independence criteria. *International Journal of Approximate Reasoning, 24*, 11–37.

de Campos, L. M., & Huete, J. F. (2000). Approximating causal orderings for Bayesian networks using genetic algorithms and simulated annealing. *In Proceedings of the Eighth IPMU Conference*, 333–340.

de Campos, L. M., & Puerta, J. M. (2001). Stochastic local and distributed search algorithms for learning belief networks. *In Proceedings of the III International Symposium on Adaptive Systems: Evolutionary Computation and Probabilistic Graphical Model*, 109–115.

de Campos, L. M., & Puerta, J. M. (2001). Stochastic local search algorithms for learning belief networks: Searching in the space of orderings. *Lecture Notes in Artificial Intelligence, 2143*, 228–239.

Dor, D., & Tarsi, M. (1992). A simple algorithm to construct a consistent extension of a partially oriented graph. Tech. rep. R-185, Cognitive Systems Laboratory, Department of Computer Science, UCLA.

Friedman, N., & Goldszmidt, M. (1996). Learning Bayesian networks with local structure. *In Proceedings of the Twelfth Conference on Uncertainty in Artificial Intelligence*, 252–262.

Friedman, N., & Koller, D. (2000). Being Bayesian about network structure. *In Proceedings of the Sixteenth Conference on Uncertainty in Artificial Intelligence*, 201–210.

Friedman, N., Nachman, I., & Peér, D. (1999). Learning Bayesian network structure from massive datasets: The "sparse candidate" algorithm. *In Proceedings of the Fifteenth Conference on Uncertainty in Artificial Intelligence*, 206–215.

Geiger, D., & Heckerman, D. (1995). A characterisation of the Dirichlet distribution with application to learning Bayesian networks. *In Proceedings of the Eleventh Conference on Uncertainty in Artificial Intelligence*, 196–207.

Geiger, D., Paz, A., & Pearl, J. (1990). Learning causal trees from dependence information. *In Proceedings of AAAI-90*, 770–776.

Geiger, D., Paz, A., & Pearl, J. (1993). Learning simple causal structures. *International Journal of Intelligent Systems, 8*, 231–247.

Gillispie, S. B., & Perlman, M. D. (2001). Enumerating Markov equivalence classes of acyclic digraphs models. *In Proceedings of the Seventeenth Conference on Uncertainty in Artificial Intelligence*, 171–177.

Glover, F. (1989). Tabu search, Part I. *ORSA Journal of Computing, 1*, 190–206.

Heckerman, D. (1996). Bayesian networks for knowledge discovery. In U.M. Fayyad, G. Piatetsky-Shapiro, P. Smyth, R. Uthurusamy (Eds.), *Advances in Knowledge Discovery and Data Mining.* Cambridge: MIT Press, 273–305.

Heckerman, D., Geiger, D., & Chickering, D. M. (1995). Learning Bayesian networks: The combination of knowledge and statistical data. *Machine Learning, 20*, 197–243.

Herskovits, E. (1991). *Computer-based probabilistic networks construction.* Ph.D thesis, Medical Information Sciences, Stanford University.

Herskovits, E., & Cooper, G. F. (1990). Kutató: An entropy-driven system for the construction of probabilistic expert systems from databases. *In Proceedings of the Sixth Conference on Uncertainty in Artificial Intelligence*, 54–62.

Huete, J. F., & de Campos, L. M. (1993). Learning causal polytrees. *Lecture Notes in Computer Science, 747*, 180–185.

Jensen, F. V. (1996). *An Introduction to Bayesian Networks.* UCL Press.

Kocka, T., & Castelo, R. (2001). Improved learning of Bayesian networks. *In Proceedings of the Seventeenth Conference on Uncertainty in Artificial Intelligence*, 269–276.

Kohavi, R., John, G., Long, R., Manley, D., & Pfleger, K. (1994). MLC++: A machine learning library in C++. *In Proceedings of the Sixth International Conference on Tools with Artificial Intelligence*, 740–743.

Kullback, S. (1968). *Information Theory and Statistics.* Dover Publication.

Lam, W., & Bacchus, F. (1994). Learning Bayesian belief networks. An approach based on the MDL principle. *Computational Intelligence, 10*, 269–293.

Larrañaga, P., Poza, M., Yurramendi, Y., Murga, R., & Kuijpers, C. (1996). Structure learning of Bayesian networks by genetic algorithms: A performance analysis of control parameters. *IEEE Transactions on Pattern Analysis and Machine Intelligence, 18*, 912–926.

Larrañaga, P., Kuijpers, C., & Murga, R. (1996). Learning Bayesian network structures by searching for the best ordering with genetic algorithms. *IEEE Transactions on System, Man and Cybernetics, 26*, 487–493.

Madigan, D., Anderson, S. A., Perlman, M. D., & Volinsky, C. T. (1996). Bayesian model averaging and model selection for Markov equivalence classes of acyclic digraphs. *Communications in Statistics – Theory and Methods, 25*, 2493–2520.

Madigan, D., & Raftery, A. (1994). Model selection and accounting for model uncertainty in graphical models using Occam's window. *Journal of the American Statistics Association, 89*, 1535–1546.

Meek, C. (1995). Causal inference and causal explanation with background knowledge. *In Proceedings of the Eleventh Conference on Uncertainty in Artificial Intelligence*, 403–410.

Muntenau, P., & Cau, D. (2000). Efficient score-based learning of equivalence classes of Bayesian networks. *Lecture Notes in Artificial Intelligence, 1910*, 96–105.

Myers, J. W., Laskey, K. B., & Levitt, T. (1999). Learning Bayesian networks from incomplete data with stochastic search algorithms. *In Proceedings of the Fifteenth Conference on Uncertainty in Artificial Intelligence*, 476–485.

Pearl, J. (1988). *Probabilistic Reasoning in Intelligent Systems: Networks of Plausible Inference*. San Mateo: Morgan Kaufmann.

Pearl, J., & Verma, T. S. (1990). Equivalence and synthesis of causal models. *In Proceedings of the Sixth Conference on Uncertainty in Artificial Intelligence*, 220–227.

Puerta, J. M. (2001). *Métodos locales y distribuidos para la construcción de redes de creencia estáticas y dinámicas* (in Spanish). Ph.D. thesis, Department of Computer Science and Artificial Intelligence, University of Granada.

Ramoni, M., & Sebastiani, P. (1997). Learning Bayesian networks from incomplete databases. *In Proceedings of the Thirteenth Conference on Uncertainty in Artificial Intelligence*, 401–408.

Rebane, G., & Pearl, J. (1987). The recovery of causal poly-trees from statistical data. In L.N. Kanal, T.S. Levitt, J.F. Lemmer (Eds.), *Uncertainty in Artificial Intelligence 3*, Amsterdam: North-Holland, 222–228.

Schwarz, G. (1978). Estimating the dimension of a model. *Annals of Statistics, 6*, 461–464.

Singh, M., & Valtorta, M. (1993). An algorithm for the construction of Bayesian network structures from data. *In Proceedings of the Ninth Conference on Uncertainty in Artificial Intelligence*, 259–265.

Singh, M., & Valtorta, M. (1995). Construction of Bayesian network structures from data: A brief survey and an efficient algorithm. *International Journal of Approximate Reasoning, 12*, 111–131.

Spirtes, P., Glymour, C., & Scheines, R. (1993). *Causation, Prediction and Search*. Lecture Notes in Statistics 81, New York: Springer Verlag.

Spirtes, P., & Meek, C. (1995). Learning Bayesian networks with discrete variables from data. *In Proceedings of the First International Conference on Knowledge Discovery and Data Mining*, 294–299.

Steck, H. (2000). On the use of skeletons when learning in Bayesian networks. *In Proceedings of the Sixteenth Conference on Uncertainty in Artificial Intelligence*, 558–565.

Suzuki, J. (1993). A construction of Bayesian networks from databases based on the MDL principle. *In Proceedings of the Ninth Conference on Uncertainty in Artificial Intelligence*, 266–273.

Suzuki, J. (1996). Learning Bayesian belief networks based on the minimum description length principle: An efficient algorithm using the B&B technique. *In Proceedings of the Thirteenth International Conference on Machine Learning*, 462–470.

Tian, J. (2000). A branch-and-bound algorithm for MDL learning Bayesian neworks. *In Proceedings of the Sixteenth Conference on Uncertainty in Artificial Intelligence*, 580–587.

Verma, T., & Pearl, J. (1990). Causal networks: Semantics and expressiveness. In R.D. Shachter, T.S. Lewitt, L.N. Kanal, J.F. Lemmer (Eds.), *Uncertainty in Artificial Intelligence, 4*, Amsterdam: North-Holland, 69–76.

Wermuth, N., & Lauritzen, S. (1983). Graphical and recursive models for contingence tables. *Biometrika, 72*, 537–552.

Wong, M. L., Lam, W., & Leung, K. S. (1999). Using evolutionay computation and minimum description length principle for data mining of probabilistic knowledge. *IEEE Transactions on Pattern Analysis and Machine Intelligence, 21*, 174–178.